\documentclass[a4paper,fleqn]{cas-sc}

\usepackage[square,numbers]{natbib}

\usepackage{tikz}
\usetikzlibrary{shapes,snakes,positioning,shapes.misc}

\usepackage[noend]{algpseudocode}
\usepackage{algorithmicx,algorithm}
\usepackage{amsmath}

\usepackage[justification=centering]{caption}
\usepackage{makecell}
\def\tsc#1{\csdef{#1}{\textsc{\lowercase{#1}}\xspace}}
\tsc{WGM}
\tsc{QE}
\tsc{EP}
\tsc{PMS}
\tsc{BEC}
\tsc{DE}
\usepackage{color}
\usepackage{booktabs}
\newcommand{\tblack}{\textcolor{black}}	
\usepackage{ulem}
\begin{document}
\let\WriteBookmarks\relax
\def\floatpagepagefraction{1}
\def\textpagefraction{.001}
\shorttitle{A Survey of Deep Visual Cross-Domain Few-Shot Learning}
\shortauthors{Lijuan Duan et~al.}

\title [mode = title]{A Survey of Deep Visual Cross-Domain Few-Shot Learning}

\author[1,2,3]{Wenjian~Wang}[orcid=0000-0002-3941-8952]
\ead{wangwj@emails.bjut.edu.cn}

\author[1,2,3]{Lijuan~Duan}
\cormark[1]
\ead{ljduan@bjut.edu.cn}

\author[4]{Yuxi~Wang}
\ead{yuxiwang93@gmail.com}

\author[4]{Junsong~Fan}
\ead{junsong.fan@ia.ac.cn}
\author[1,2,3]{Zhi~Gong}
\ead{gongzhi97@emails.bjut.edu.cn}

\author[4,5]{Zhaoxiang~Zhang}
\ead{zhaoxiang.zhang@ia.ac.cn}
\cormark[1]

\credit{Conceptualization of this study, Methodology, Software}

\address[1]{Faculty of Information Technology, Beijing University  of Technology, China}
\address[2]{Beijing Key Laboratory of Trusted Computing, Beijing, China}
\address[3]{National Engineering Laboratory for Key Technologies of Information Security Level Protection, Beijing, China}

\address[4]{Centre for Artificial Intelligence and Robotics, (HKISI\_CAS)}

\address[5]{Institute of Automation, Chinese Academy of Sciences (NLPR, CASIA, UCAS)}

\credit{Data curation, Writing - Original draft preparation}

\begin{abstract}
Few-Shot transfer learning has become a major focus of research as it allows recognition of new classes with limited labeled data. While it is assumed that train and test data have the same data distribution, this is often not the case in real-world applications. This leads to decreased model transfer effects when the new class distribution differs significantly from the learned classes. Research into Cross-Domain Few-Shot (CDFS) has emerged to address this issue, forming a more challenging and realistic setting. In this survey, we provide a detailed taxonomy of CDFS from the problem setting and corresponding solutions view. We summarise the existing CDFS network architectures and discuss the solution ideas for each direction the taxonomy indicates.
Furthermore, we introduce various CDFS downstream applications and outline classification, detection, and segmentation benchmarks and corresponding standards for evaluation. We also discuss the challenges of CDFS research and explore potential directions for future investigation. Through this review, we aim to provide comprehensive guidance on CDFS research, enabling researchers to gain insight into the state-of-the-art while allowing them to build upon existing solutions to develop their own CDFS models.
\end{abstract}

%


\maketitle
%
%
\section{INTRODUCTION}
Strong computing power has boosted the development of deep learning, thereby aiding computer vision in making considerable progress. Research fields such as image classification, object detection, and semantic segmentation continue to shape the development of computer vision in innovative ways. Current deep models require a large volume of annotated data for training, but this data is typically expensive and labor-intensive. Few labeled data are in certain fields to enable the model to recognize new categories, triggering the need for few-shot learning \cite{Snell2017PrototypicalNF,Sung2017LearningTC,Vinyals2016MatchingNF}. Although few-shot learning assumes that the training and testing data come from the same domain, domain shift is commonly seen in real-world scenarios \cite{Wang2022TPSNTM}. In these contexts, cross-domain few-shot learning (CDFS) offers a promising solution to the few-shot learning problem by simultaneously addressing both domain shift and data scarcity \cite{Wang2022RememberTD,Guo2019ABS,Tseng2020CrossDomainFC}.  

This paper reviews and categorizes the current research in CDFS based on problem sets, solutions, and applications. Regarding problem setting, we identify two main models for CDFS research with multiple data sources: single- and multiple-model (where the latter involves training a model for each source and subsequently aggregating them). For single-source modeling, sub-categorizations are achieved based on whether the target domain data is accessible (through supervised/unsupervised means) or forbidden. We also comprehensively introduce the popular CDFS classification benchmark datasets and highlight the unified transfer effects across datasets.

Regarding solution, we cover four main approaches: Image Augmentation, Feature Augmentation, Decoupling, and Fine-tuning. Image and Feature Augmentation rely on the mutual image/feature transformation to enrich data distribution. At the same time, Decoupling Based Methods distinguish domain-irrelevant from domain-specific features, and Fine-tuning uses meta-learning and distribution alignment to extract transferable features. We further discuss its application to various fields and identify potential future research directions.

The contributions of this article are summarized as follows:
	\begin{enumerate}
		\item We summarized the cross-domain research in few-shot learning. The existing solutions are summarized in detail according to the different problem settings.
		\item For different CDFS solutions, we give a detailed sorting and classification from features, images, and network architecture. We also summarize the application of CDFS in different scenarios.
		\item We gave the benchmark of the current CDFS, providing a unified reference precision for subsequent research. We provide the future direction of CDFS in terms of challenges, technical solutions, and applications.
	\end{enumerate}

\begin{figure}[t]
		\centering
		\includegraphics[width=\linewidth,trim={0cm 0cm 0cm 0cm}, clip]{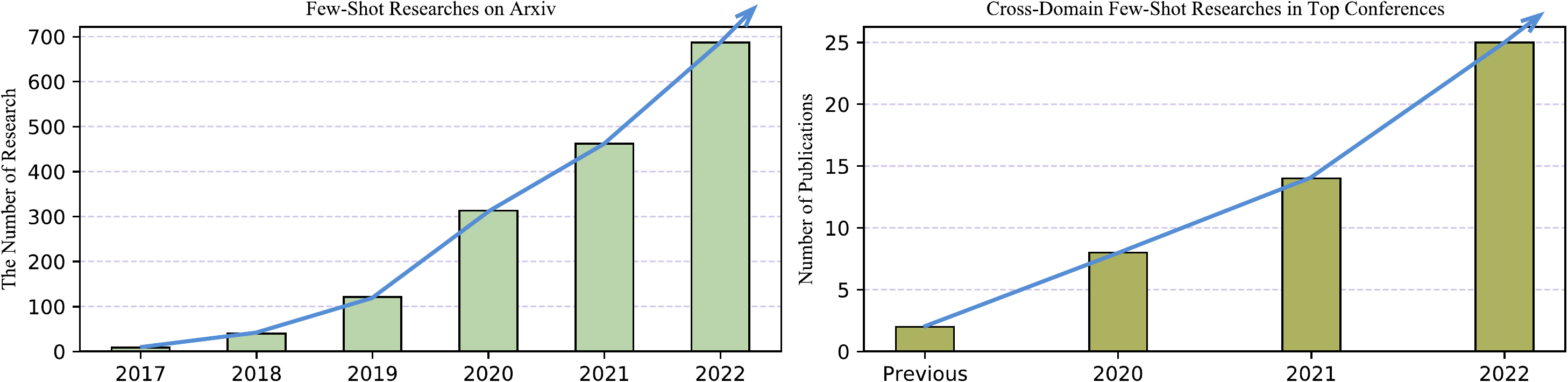}
    		\caption{The related researches on few-shot learning increased from 2017 to 2022.
		}
		\vspace{-0.3cm}
		\label{fig:overall} 
	\end{figure}
 
\section{CONCEPTS AND PRELIMINARIES}
\subsection{Few-Shot Learning}
Few-Shot Learning (FSL) is a transfer learning technique designed to learn novel classes from limited annotated labels \cite{Wang2019GeneralizingFA}. In FSL, two assumptions must be met to ensure its successful application. Firstly, the classes between the train and finetune process must be distinct; there must be no class intersection between the two processes, i.e., $C_{b}\cap C_{n} = \varnothing$. Secondly, the annotation labels for each class in the finetune process must be limited (or even only have one annotation label). 
Specifically, there are two datasets $D_{b}$ and $D_{n}$, where the classes $C_{b} \in D_{b}$ have sufficient labels and $C_{n} \in D_{n}$ only have limited labels. The FSL model aims to train on dataset $D_{b}$, then apply a few annotation labels to finetune novel classes of dataset $D_{n}$ during the test process.

\subsection{Few-Shot Domain Adaptation}
Few-Shot Domain Adaptation (FSDA) is a powerful combination of Few-Shot and Domain Adaptation, making Few-Shot learning more difficult. Unlike standard Few-Shot Learning, which assumes a shared data distribution between the training dataset $D_b$ and finetuning dataset $D_n$, FSDA identifies a domain gap between the two datasets. Two kinds of FSDA: the supervised kind, explored in \cite{Motiian2017FewShotAD}, and the unsupervised kind, explored in \cite{Yue2021PrototypicalCS}. This research seeks to bridge the difference between the two domains and fully use the limited annotations available. The goal of FSDA is thus to align two data domains while utilizing the limited annotation data.

\subsection{Cross-Domain Few-Shot}
Cross-domain few-shot (CDFS) learning, a recently emerged field in few-shot learning, assumes there is a significant domain gap between source and target domains, making the task more challenging compared to the traditional few-shot domain adaptation (FSDA) approach \cite{Wang2022RememberTD}. As shown in Fig. \ref{fig:overall}, the interest in CDFS research has grown steadily from 2020 to 2022. The following sections provide a detailed overview of the problem sets, solutions, and applications of CDFS. Section \ref{sec:SETTING} summarizes the problem divisions of CDFS, Section \ref{sec:SOLUTION} reviews the different solutions proposed for CDFS, Section \ref{BENCHMARK} provides the benchmark, Section \ref{sec:Application} provides various applications of CDFS, and Section \ref{sec:FUTURE} outlines directions for future research.

\subsection{Domain Adaptation \& Generalization}
Domain Adaptation (DA) requires a model to be trained on a source domain and tested on a target domain with different data distribution. This can be done in a supervised \cite{Saito2019SemiSupervisedDA,Chen2021SemisupervisedDA} or unsupervised manner \cite{Liang2020SourceDU,Xu2021CDTransCT}, assuming the same classes are present in both domains. DA focuses on aligning the distributions of two domains \cite{Liu2021AdversarialUD}. In an unsupervised setting, the source model trains the feature extractor using the data of the target domain, thus accomplishing domain alignment. In contrast, Domain Generalization (DG) is more complex and valuable than DA \cite{Wang2021GeneralizingTU}, as the target domain data is not accessible during training \cite{Zhou2021DomainGA}. After completing source domain training, the model must be tested directly on the target domain.

\subsection{Meta-Learning}
Humans constantly summarize experiences during the learning process. When faced with a novel environment, humans can draw upon acquired knowledge to adapt quickly to new tasks. Meta-Learning, or "learning to learn," is an approach that collects experiences across various tasks to provide valuable information for transfer learning \cite{Thrun1998LearningTL}. As a form of meta-learning, task learning encompasses various applications, including classification problems, regression, and mixed tasks \cite{Sun2019MetaTransferLT,Achille2019Task2VecTE,Vuorio2019MultimodalMM}. Meta-learning aims to distill knowledge, which can be manifested in various network architectures. For example, Reptile \cite{Nichol2018ReptileAS}, a model-agnostic meta-learning algorithm, utilizes the prediction of model parameters as its knowledge representation. Alternatively, DMML \cite{Chen2019DeepMM} and OFMA \cite{Nichol2018OnFM} encode knowledge via an embedding function and an initialization parameter, respectively.

\begin{table}[h]
 \footnotesize
\tabcolsep=3pt
\caption{The different related researches.
DA and DG transfer source knowledge to target domain, but DA have the accessible of target data.
FSL assume that only have limited fine-tuning data and have no class intersection between training and testing data.
CDFS and FSDA consider the few-shot learning and domain shift together, but FSDA have the target domain accessible and have the same domain class.
 }
 \centering
\begin{tabular}{l|lllll} 
\toprule
\multicolumn{1}{c|}{Research} & \multicolumn{1}{c}{Training Domain} & \multicolumn{1}{c}{Test Domain} & \multicolumn{1}{c}{Domain Class Intersection} & \multicolumn{1}{c}{Fine-tuning Size} & \multicolumn{1}{c}{Test Data Training}  \\ 
\hline
Cross-Domain Few-Shot         & $D_s$                                  & $D_t$                              & No                                            & Few                                  & Forbidden                               \\
Few-Shot Domain Adaptation    & $D_s$                                  & $D_t$                              & Same                                          & Few                                  & Accessible                              \\
Few-Shot Learning             & $D_s$                                  & $D_s$                              & No                                            & Few                                  & Inexistent                              \\
Domain Adaptation             & $D_s$                                  & $D_t$                              & Same                                          & Sufficient                           & Accessible                              \\
Domain Generalization         & $D_s$                                  & $D_t$                              & Same                                          & Inexistent                           & Forbidden                               \\
\bottomrule
\end{tabular}
\end{table}

 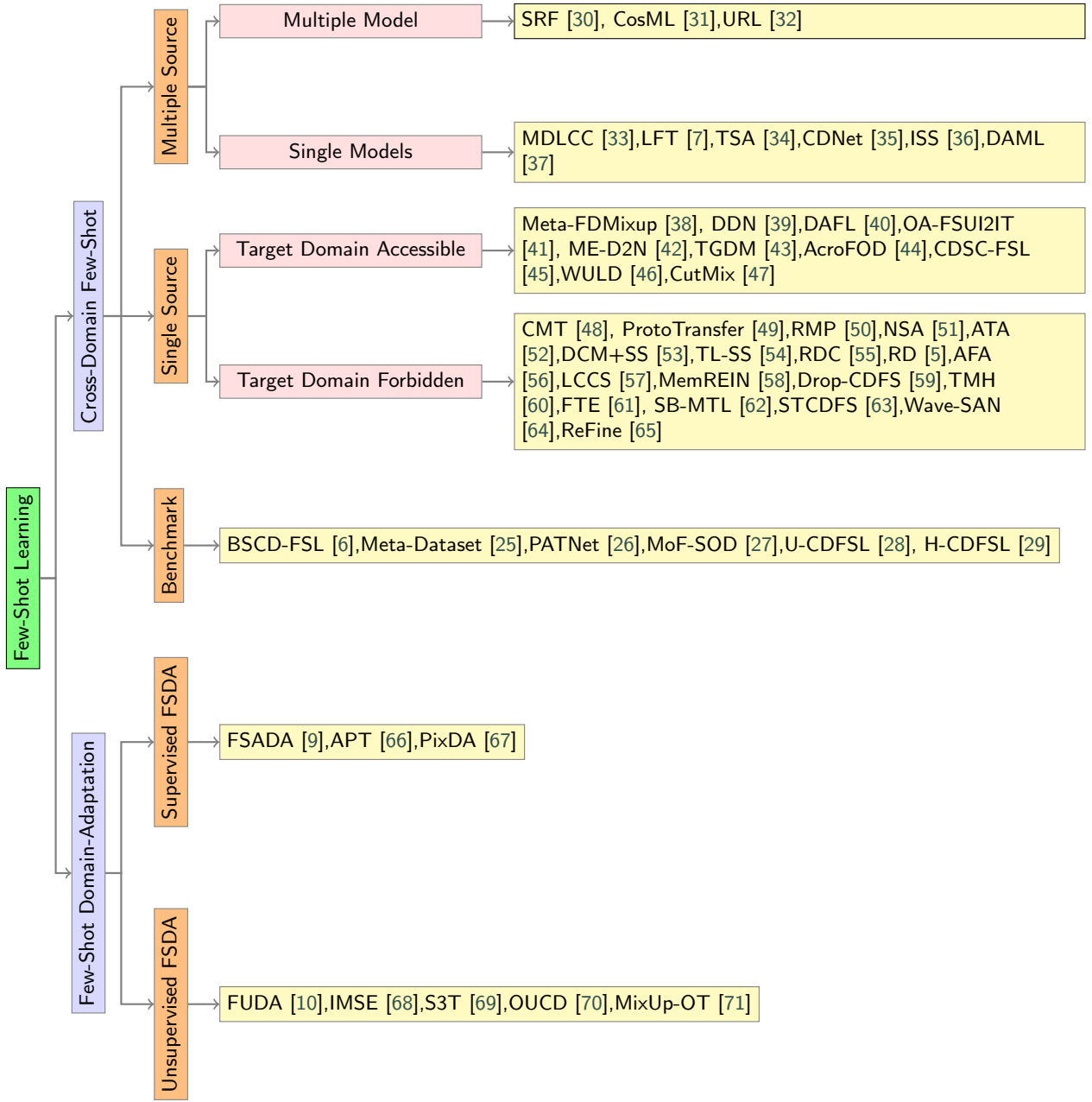
\begin{figure}
\begin{center}
\begin{tikzpicture}
  \node at (-10,-5.5) [rectangle, draw=black,fill=green!50](a){\rotatebox{90}{\tblack{Few-Shot Learning}}};
  
  \node at (-9,-1.5) [rectangle,draw,fill=blue!15](b){\rotatebox{90}{\tblack{Cross-Domain Few-Shot}}};
  \node at (-9,-10) [rectangle,draw,fill=blue!15](c){\rotatebox{90}{\tblack{Few-Shot Domain-Adaptation}}};
  
  \node[anchor=west] at (-8,2) [rectangle,draw,fill=orange!50](d){\rotatebox{90}{\tblack{Multiple Source}}};
  \node[anchor=west] at (-8,-1.5) [rectangle,draw,fill=orange!50](e){\rotatebox{90}{\tblack{Single Source}}};
  \node[anchor=west] at (-8,-5) [rectangle,draw,fill=orange!50](f){\rotatebox{90}{\tblack{Benchmark}}};
  \node[anchor=west] at (-8,-8) [rectangle,draw,fill=orange!50](g){\rotatebox{90}{\tblack{Supervised FSDA}}};
  \node[anchor=west] at (-8,-12) [rectangle,draw,fill=orange!50](h){\rotatebox{90}{\tblack{Unsupervised FSDA}}};
  
  \node[anchor=west] at (-7,3) [rectangle,draw,fill=pink!50, minimum width=4cm](i){\tblack{Multiple Model}};
  \node[anchor=west] at (-7,1) [rectangle,draw,fill=pink!50, minimum width=4cm](j){\tblack{Single Models}};
  \node[anchor=west] at (-7,-0.5) [rectangle,draw,fill=pink!50, minimum width=4cm](k){\tblack{Target Domain Accessible}};
  \node[anchor=west] at (-7,-2.5) [rectangle,draw,fill=pink!50, minimum width=4cm](l){\tblack{Target Domain Forbidden}};

    \node[anchor=west] at (-7,-5) [rectangle,draw, fill=yellow!30, minimum width=4cm](q){\tblack{BSCD-FSL \cite{Guo2019ABS},Meta-Dataset \cite{Triantafillou2019MetaDatasetAD},PATNet \cite{Lei2022CrossDomainFS},MoF-SOD \cite{Lee2022RethinkingFO},U-CDFSL \cite{Oh2022UnderstandingCF}, H-CDFSL \cite{Zhang2022HowWD}}};

  \draw[->,thick] (a.east)--(-9.5,-5.5)--(-9.5,-1.5)--(b.west);
  \draw[->,thick] (a.east)--(-9.5,-5.5)--(-9.5,-10)--(c.west);

    \draw[->,thick] (f.east)--(q.west);

  \draw[->,thick] (c.east)--(-8.5,-10)--(-8.5,-8)--(g.west);
  \draw[->,thick] (c.east)--(-8.5,-10)--(-8.5,-12)--(h.west);

  \draw[->,thick] (b.east)--(-8.5,-1.5)--(-8.5,2)--(d.west);
  \draw[->,thick] (b.east)--(-8.5,-1.5)--(e.west);
  \draw[->,thick] (b.east)--(-8.5,-1.5)--(-8.5,-5)--(f.west);

  \draw[->,thick] (d.east)--(-7.2,2)--(-7.2,3)--(i.west);
  \draw[->,thick] (d.east)--(-7.2,2)--(-7.2,1)--(j.west);

  \draw[->,thick] (e.east)--(-7.2,-1.5)--(-7.2,-0.5)--(k.west);
  \draw[->,thick] (e.east)--(-7.2,-1.5)--(-7.2,-2.5)--(l.west);

    \node[anchor=west] at (-2.5,3) [rectangle,draw=black, fill=yellow!30, text width=8.5cm](n){\tblack{SRF \cite{Dvornik2020SelectingRF}, CosML \cite{Peng2020CombiningDM},URL \cite{Li2021UniversalRL}}};
  \node[anchor=west] at (-2.5,1) [rectangle,draw, fill=yellow!30, text width=8.5cm](m){\tblack{MDLCC \cite{Xiao2020MultiDomainLF},LFT \cite{Tseng2020CrossDomainFC},TSA \cite{Li2022CrossdomainFL},CDNet \cite{Zou2022LearntoDecomposeCD},ISS \cite{Xu2022CrossDomainFC},DAML \cite{Lee2022DomainAgnosticMF}}};
    \node[anchor=west] at (-2.5,-0.5) [rectangle,draw, fill=yellow!30, text width=8.5cm](o){\tblack{Meta-FDMixup \cite{fu2021meta}, DDN \cite{Islam2021DynamicDN},DAFL \cite{Zhao2020DomainAdaptiveFL},OA-FSUI2IT \cite{Zhao2022OAFSUI2ITAN}, ME-D2N \cite{Fu2022MED2NMD},TGDM \cite{Zhuo2022TGDMTG},AcroFOD \cite{Gao2022AcroFODAA},CDSC-FSL \cite{Chen2022CrossDomainCF},WULD \cite{Yao2021CrossdomainFL},CutMix \cite{Nakamura2022FewshotAO}}};
    \node[anchor=west] at (-2.5,-2.5) [rectangle,draw, fill=yellow!30, text width=8.5cm](p){\tblack{CMT \cite{Teshima2020FewshotDA}, ProtoTransfer \cite{Medina2020SelfSupervisedPT},RMP \cite{Zou2020RevisitingMP},NSA \cite{Liang2021BoostingTG},ATA \cite{Wang2021CrossDomainFC},DCM+SS \cite{Tao2022PoweringFI},TL-SS \cite{Yuan2022TaskLevelSF},RDC \cite{Li2021RankingDC},RD \cite{Wang2022RememberTD},AFA \cite{Hu2022AdversarialFA},LCCS \cite{Zhang2022FewShotAO},MemREIN \cite{Xu2022MemREINRT},Drop-CDFS \cite{Tu2021ADS},TMH \cite{Jiang2020ATM},FTE \cite{Liu2020FeatureTE}, SB-MTL \cite{Cai2020SBMTLSM},STCDFS \cite{Liu2021SelfTaughtCF},Wave-SAN \cite{Fu2022WaveSANWB},ReFine \cite{Oh2022ReFineRB}}};

    \draw[->,thick] (i.east)--(n.west);
    \draw[->,thick] (j.east)--(m.west);
    \draw[->,thick] (k.east)--(o.west);
    \draw[->,thick] (l.east)--(p.west);

    \node[anchor=west] at (-7,-8) [rectangle,draw, fill=yellow!30, minimum width=4cm](r){\tblack{FSADA \cite{Motiian2017FewShotAD},APT \cite{Li2021FewShotDA},PixDA \cite{Tavera2021PixelbyPixelCA}}};

    \node[anchor=west] at (-7,-12) [rectangle,draw, fill=yellow!30, minimum width=4cm](s){\tblack{FUDA \cite{Yue2021PrototypicalCS},IMSE \cite{Huang2021FewshotUD},S3T \cite{Phoo2020SelftrainingFF},OUCD \cite{DInnocente2020OneShotUC},MixUp-OT \cite{Xu2022FewShotDA}}};

    \draw[->,thick] (g.east)--(r.west);
    \draw[->,thick] (h.east)--(s.west);

\end{tikzpicture}
\caption{Taxonomy of CDFS.
For domain researches of CDFS, we split the Cross-Domain Few-Shot into Multiple Source and Single Source setting.
For Multiple Source, we divide the research into Multiple Models and Single Model according to the models used. For Single Source, we divide the research into Target Domain Accessible and Target Domain Forbidden according to whether the target domain can access.
}
\end{center}
\label{fig:Taxonomy}
\vspace{-1cm}
\end{figure}

\section{DIFFERENT SETTING OF CDFS}\label{sec:SETTING}
\begin{figure}[h]
		\centering
		\includegraphics[width=0.9\textwidth,trim={0cm 0cm 0cm 0cm}, clip]{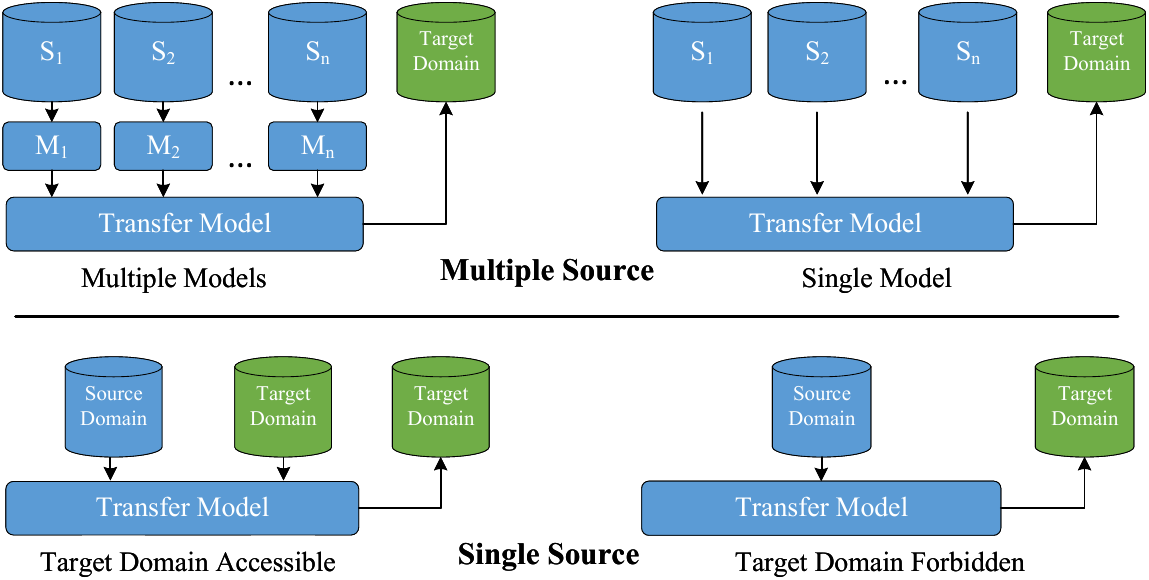}
    		\caption{Different setting of CDFS.
      For the CDFS research, there are Multiple Source and Single Source setting. 
		}
		\vspace{-0.3cm}
		\label{fig:setting} 
	\end{figure}
 
We group recent CDFS research into three categories: \textbf{Multiple Source}, \textbf{Single Source}, and \textbf{Benchmark}. Multiple Source research includes two solutions: multiple models, and a single model. Single Source can be divided into two subcategories, depending on whether the target domain is accessible or forbidden, as depicted in Fig. \ref{fig:setting}. We have also provided a summary of benchmarks for CDFS in various research fields.

\subsection{Multiple Source CDFS}
Different domains can provide essential data distributions for deep model learning, making it intuitive to collect multiple source domain data to address the CDFS problem. Specifically, researchers first collect several source domain datasets with disjoint classes. They then employ specific training strategies to capture domain information and combine the results into a final transfer model. Typically, these domains span various scenarios, such as CropDisease \cite{Mohanty2016UsingDL}, EuroSAT \cite{Helber2019EuroSATAN}, and ChestX \cite{Wang2017ChestXRay8HC}. The learning strategies of Multiple Source CDFS can be divided into \textbf{Multiple-Model} and \textbf{Single-Model} approaches, as depicted in Fig. \ref{fig:setting}.

\subsubsection{Multiple-Model For Multiple Source} 
The domain will have its specific model for the Multiple-Model Methods. The learning process contains the training and aggregation stage. Each domain optimizes the domain-specific model in the training stage, then combines all the models into a single model, which will serve as the final transfer model in the aggregation stage, as shown in Fig. \ref{fig:setting} left part.
The training stage gives the model diverse feature distribution or highly adaptable parameters. The subsequent aggregation stage focuses on feature reuse, which aims to achieve the best transfer effect.

	\begin{figure}[h]
        \centering
		\includegraphics[width=0.9\textwidth,trim={0cm 0cm 0cm 0cm}, clip]{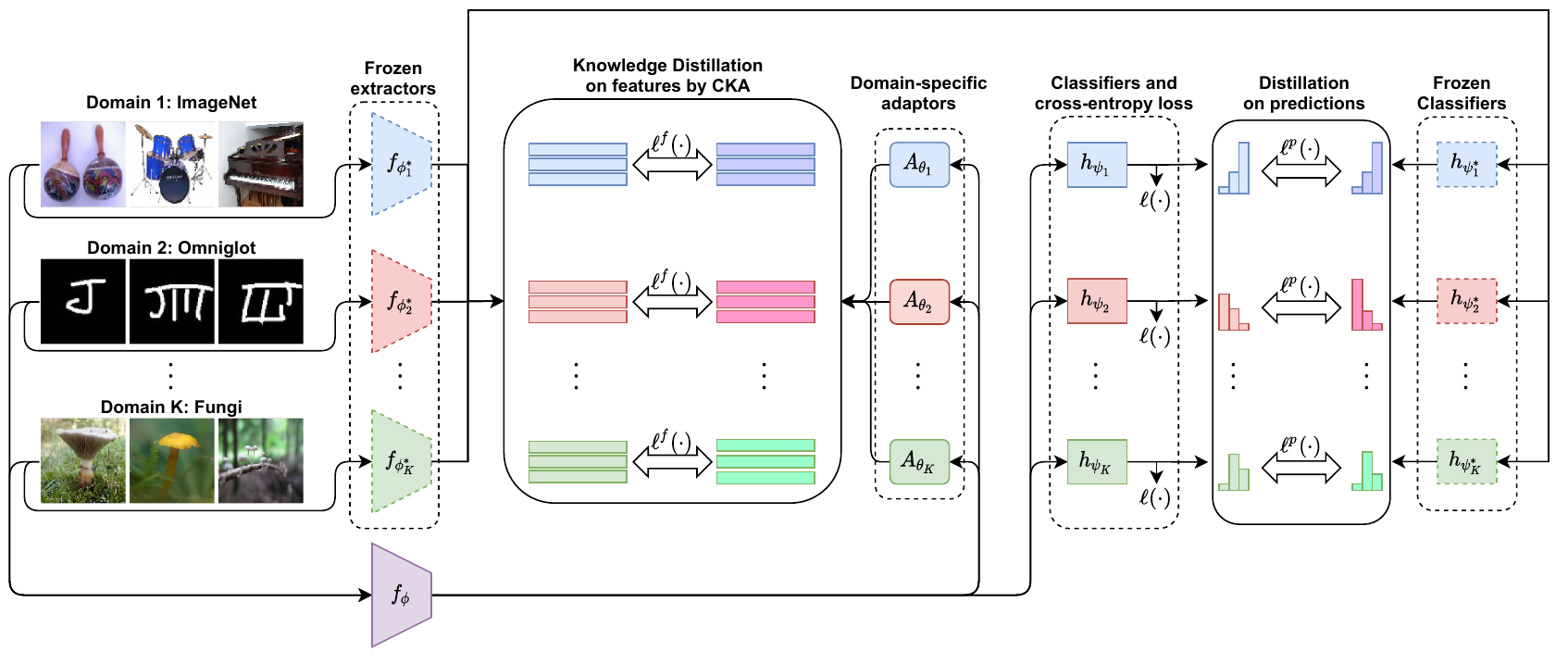}
    		\caption{The URL \cite{Li2021UniversalRL} use the knowledge distillation to collect multiple source information.
		}
		\vspace{-0.3cm}
		\label{fig:url}
	\end{figure}
The URL \cite{Li2021UniversalRL} is a typical model based on Multiple-Model Methods, as shown the Fig. \ref{fig:url}.
The framework first trains $K$ teacher network $\{f_{\phi_{1}^{*}}, f_{\phi_{2}^{*}}...f_{\phi_{K}^{*}}\}$ for $K$ domains on the training stage, then use the student network $f_{\phi}$ to distill the domain information on the aggregation stage. 
The student network $f_{\phi}$ can observe all the source domain datasets, but the label comes from the teacher networks' output.
URL \cite{Li2021UniversalRL} focuses on aggregation learning and aims to learn a single set of universal
representations. For the aggregation, use the Domain-specific adaptors $\{A_{\theta_1},A_{\theta_2}...A_{\theta_K}\}$ and Classifiers $\{h_{\psi_1},h_{\psi_2}...h_{\psi_K}\}$ to match the teacher networks' output.
URL \cite{Li2021UniversalRL} has a fixed computational cost regardless of the number of domains at inference unlike them.

\begin{figure}[h]
		\centering
		\includegraphics[width=0.9\textwidth,trim={0cm 0cm 0cm 0cm}, clip]{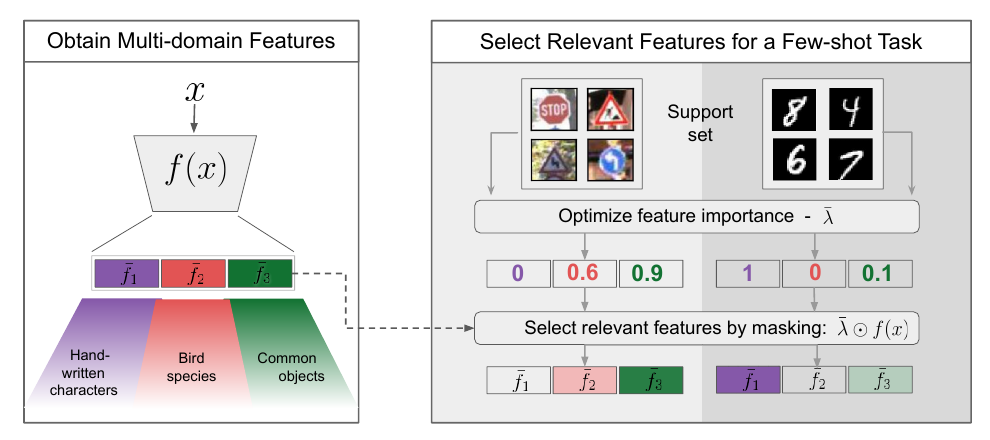}
  \caption{The SRF \cite{Dvornik2020SelectingRF} focus on the more relevant domain feature, use multi-domain feature bank to automatically select the most relevant representations.
		}
		\vspace{-0.3cm}
		\label{fig:srf}
	\end{figure}

 SRF \cite{Dvornik2020SelectingRF} consider that more relevant domain features should be concerned, and irrelevant domain should be inhibited on the aggregation stage.
 The framework uses a set of learnable parameters $\lambda$ to select the source domain feature.
 First, SRF \cite{Dvornik2020SelectingRF} train a set of $K$ feature extractors and obtain a multi-domain feature representation, consisting of feature blocks with different semantics, as shown Fig. \ref{fig:srf} left part. 
 Then given a few-shot task, select only the relevant feature blocks from the multi-domain representation by optimizing masking parameters $\lambda$ on the support set.
 SRF \cite{Dvornik2020SelectingRF} shows that a simple feature selection mechanism can replace feature adaptation.

\begin{figure}
		\centering
		\includegraphics[height=0.4\textwidth,trim={0cm 0cm 0cm 0cm}, clip]{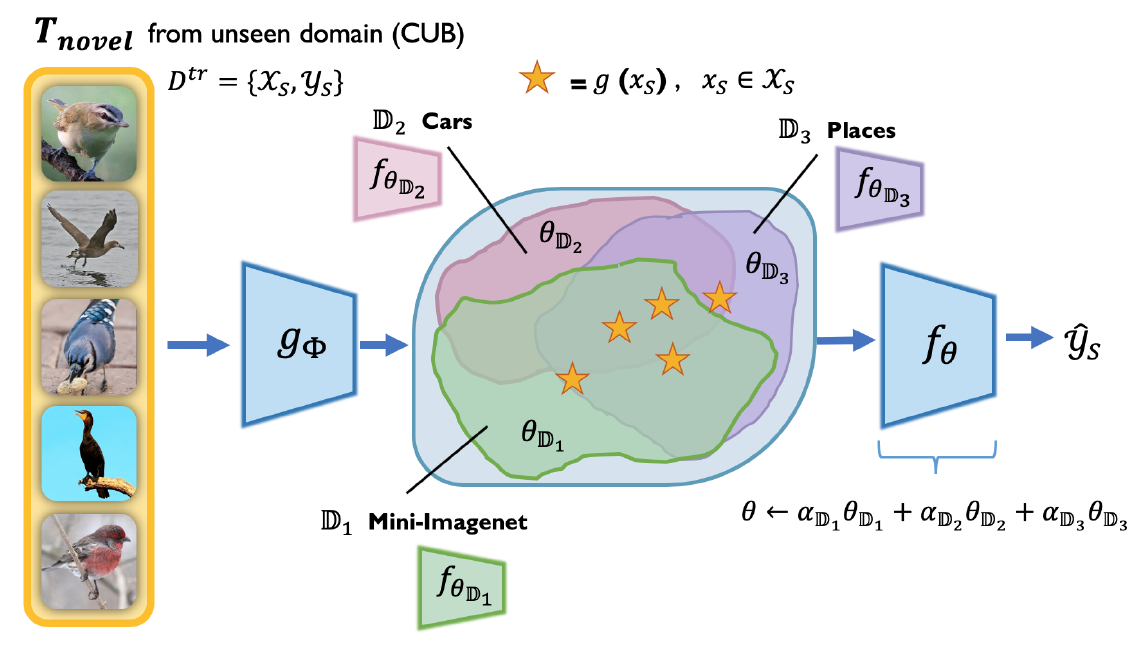}
    		\caption{The CosML \cite{Peng2020CombiningDM} first trains a set of meta-learners, one for each training domain, to learn prior knowledge (i.e., meta-parameters) specific to each domain. The
domain-specific meta-learners are then combined in the parameter space, by taking a weighted average of their meta-parameters, which is used as the initialization parameters of a task network that is quickly adapted to novel few-shot classification tasks in an unseen domain.
		}
		\vspace{-0.3cm}
		\label{fig:CosML}
	\end{figure}
Compared with feature level transfer, CosML \cite{Peng2020CombiningDM} focuses on the parameter aggregation, as shown in Fig. \ref{fig:CosML}.
For CosML \cite{Peng2020CombiningDM}, it first get parameters $\theta_{D_{k}}$ for each domain. $\theta_{D_{k}}$ have a certain transfer ability to the new domain, but they lack communication.
For the final model parameter $\theta$, CosML \cite{Peng2020CombiningDM} combine all domain parameters $\theta_{D_{k}}$ with a domain weight $\alpha_{D_{k}}$ which can reflect the domain's importance, with Eq. \ref{eq:CosML}. 
	\begin{equation}
		\begin{split}
			 \theta = \alpha_{D_{1}} \theta_{D_{1}} + \alpha_{D_{2}} \theta_{D_{2}} + ... + \alpha_{D_{k}} \theta_{D_{k}} 
		\end{split}
		\label{eq:CosML}
	\end{equation}

\subsubsection{Single-Model For Multiple Source} 
Single-Model Methods apply a unified model to learn the multiple domain feature. The Single-Model Methods can train all the source domain data, as shown in Fig. \ref{fig:setting} right part. Compared with Multiple-Models Methods, the Single-Model Methods have fewer calculation and omit the multimodel's aggregation stage. For getting high adaptability parameters, DAML \cite{Lee2022DomainAgnosticMF} focuses on the optimization strategy and learns to adapt the model to novel classes in both seen and unseen domains by data sampled from multiple domains.

LFT \cite{Tseng2020CrossDomainFC} and TSA \cite{Li2022CrossdomainFL} propose to attach feature transformation or adapters directly to a pre-trained model. The multiple source domain collectively trains these modules, improving the model's transfer ability.

\begin{figure}
		\centering
		\includegraphics[width=0.9\textwidth,trim={0cm 0cm 0cm 0cm}, clip]{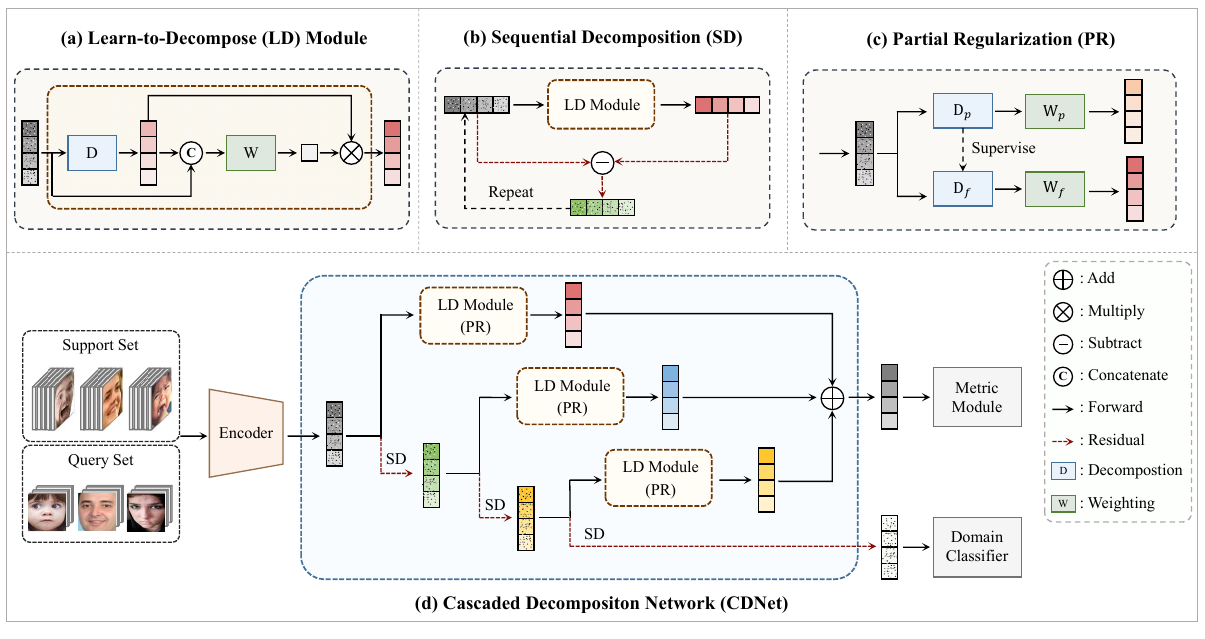}
    		\caption{The CDNet \cite{Zou2022LearntoDecomposeCD} learns the ability of learn-to-decompose that can be easily adapted to identify unseen compound expressions.
		}
		\vspace{-0.3cm}
		\label{fig:cdn}
	\end{figure}

The domain-specific information may be confused as the single model faces more than one source domain. To relieve this problem, CDNet \cite{Zou2022LearntoDecomposeCD} decomposes domain information from a given feature and extracts a domain-independent expression, as shown in the Fig. \ref{fig:cdn}.
CDNet \cite{Zou2022LearntoDecomposeCD} stacks serial LD (Learn-to-Decompose) module, which output the class feature, and SD (Sequential Decomposition) module to decompose the domain-agnostic and domain-specific feature.
Similarly, MDLCC \cite{Xiao2020MultiDomainLF} through the channel re-weighting module to decompose the different domain features. After the training, the model only needs to finetune the re-weighting module with fewer parameter changes to adapt to the new domain.

To relieve the domain labels' cost, ISS \cite{Xu2022CrossDomainFC} uses one source domain with labels and several other unlabeled domain datasets to train a single model.
The labeled domain input will be mix-enhanced with other unlabeled domain data. This can expand the labeled domain data distribution with other domains' stylization.
\subsubsection{Discussion and Summary}
Multi-source cross-domain few-shot learning uses multi-source data to improve the model's scalability. In order to optimize the learning process, the decision should analyze the efficiency, accuracy, and scalability. To this end, there are two different methods: the multi-model method (which requires communication between sources) and the single-model method (which directly processes source data and is easier to set). By appropriately using these two methods and putting enough energy into data collection, we can significantly improve the results of multi-source cross-domain few-shot learning. In addition, structured data sources, collection, and processing methods can reduce the labor-intensive nature of multi-source few-shot learning and may significantly improve the results.
 
\subsection{Single Source CDFS}
The standard few-shot learning setting requires limited annotation, and thus the use of multiple domains to train a model is not always feasible. Seeking a more realistic and manageable solution, a single source cross-domain few-shot (CDFS) method has been proposed, which uses just one source domain to train the model. This single source CDFS includes two settings: \textbf{Target Domain Accessible} and \textbf{Target Domain Forbidden}.

The standard few-shot learning setting requires limited annotation and thus makes it difficult to employ multiple domains to train a model. A single-source  method can be used to provide a more realistic and manageable solution, which uses only one source domain to train the model. This single-source CDFS includes two settings: \textbf{Target Domain Accessible} and \textbf{Target Domain Forbidden}. In the former, the target domain is available during training time, which allows the model adapts to the target domain. The target domain remains inaccessible in the latter, so the proposed models must be able to generalize across domains.

\subsubsection{Target Domain Accessible For Single Source} 
For the Single Source CDFS, using just one domain data to solve the cross-domain and few-shot problem is challenging. To address this, Target Domain Accessible methods assume that some target domain data can be accessed either via supervised learning methods \cite{Zhao2020DomainAdaptiveFL,Fu2022MED2NMD,Zhuo2022TGDMTG,Gao2022AcroFODAA} or unsupervised ones \cite{Nakamura2022FewshotAO,fu2021meta,Fu2022GeneralizedMC,Zhao2022OAFSUI2ITAN,Islam2021DynamicDN,Yao2021CrossdomainFL}. 

\textbf{Target Supervised Methods} draw on a few target domain labeled data for model training. However, the dataset's size is limited, resulting in an overfitting risk. Conversely, over-adaption and misleading can be caused by overly amplified target samples. To address this, AcroFOD \cite{Gao2022AcroFODAA} utilizes an adaptive optimization strategy that selects augmented data that is more similar to target samples rather than simply increasing the data amount. ME-D2N \cite{Fu2022MED2NMD} proposes a decompose module for knowledge distillation between two networks as domain experts to combat this.
\begin{figure}
		\centering
		\includegraphics[height=0.4\textwidth,trim={0cm 0cm 0cm 0cm}, clip]{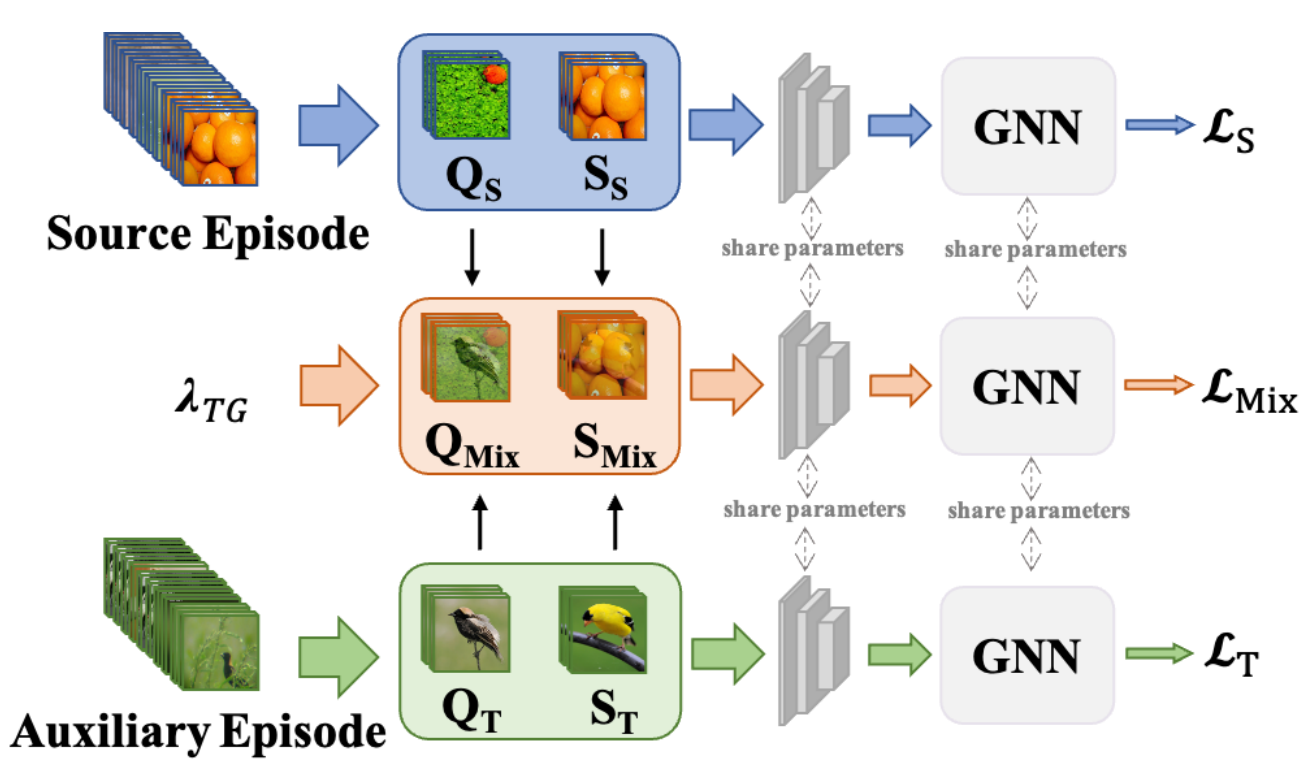}
    		\caption{The TGDM \cite{Zhuo2022TGDMTG} mix framework. Use the mixed intermediate domain reduce the domain gap.
		}
		\vspace{-0.3cm}
		\label{fig:TGDM}
	\end{figure}
The domain gap is a critical issue for CDFS, so TGDM \cite{Zhuo2022TGDMTG} design an intermediate domain generated by mixing images in the source and the target domain, as shown in the Fig. \ref{fig:TGDM}. 
The model calculates $L_{S}$, $L_{T}$, and $L_{Mix}$ three loss focus on the source, target, and intermediate domain optimized. Experiments show that the intermediate domain can effectively relieve the domain gap influence.

\textbf{Target Unsupervised Methods} are often favored due to having sufficient target domain data compared to supervised methods, but data annotation can be lacking. Meta-FDMixup \cite{fu2021meta}, and Generalized Meta-FDMixup \cite{Fu2022GeneralizedMC} use the target domain's unlabeled data to mix it with source training data. This provides class and domain labels and requires the model to learn domain-irrelevant and domain-specific features. OA-FSUI2IT \cite{Zhao2022OAFSUI2ITAN} utilizes its target's unlabeled data to transform source data and generate content consistent with the source while leaving the style to match the target--to train the network subsequently. DDN \cite{Islam2021DynamicDN} is yet another approach, in which a Dynamic Distillation Network is trained on target unlabeled data, approximate predictions from weakly-augmented versions of the same images from a teacher network are used to impose consistency regularization that matches strongly augmented versions of those same images from a student. Lastly, CDSC-FSL \cite{Chen2022CrossDomainCF} proposes a new setting focusing on the domain gap between support and query set. Here, contrast learning is utilized to align the same classes from two domains; however, it is important to note that CDSC-FSL \cite{Chen2022CrossDomainCF} requires source and target domains to have the same classes. Improvements to existing target unsupervised methods are therefore necessary to enhance transfer accuracy and effectively address any domain disparities.

\subsubsection{Target Domain Forbidden For Single Source}
Target Domain Forbidden CDFS is the most challenging cross-domain few-shot learning task, which assumes that only a single source domain is available for training and the target domain data is forbidden. Recent research \cite{Teshima2020FewshotDA,Zou2020RevisitingMP,Liang2021BoostingTG,Li2021RankingDC,Wang2021CrossDomainFC,Tu2021ADS,Jiang2020ATM,Liu2020FeatureTE,Cai2020SBMTLSM,Fu2022WaveSANWB,Oh2022ReFineRB} aims to develop networks with better adaptability by training with just one source domain. Two main strategies have been proposed: Normalization Methods and Self-Supervised Methods.

\textbf{Normalization Methods} mainly aim to improve model generation by introducing various normalization techniques to reduce the impact of domain differences. For instance, DCM+SS \cite{Tao2022PoweringFI} uses normalization to correct the bias caused by the large difference between the source and the target domain. AFA \cite{Hu2022AdversarialFA} introduces adversarial learning to improve feature diversity with normalization. LCCS \cite{Zhang2022FewShotAO} proposes to learn the parameters of BatchNorm from the source domain to mitigate the domain gap. RD \cite{Wang2022RememberTD} uses normalization to mix instance features and a learnable memory to transfer the source domain information to the target domain. MemREIN \cite{Xu2022MemREINRT} combines instance normalization with a Memory Bank to recover discriminant features. 

\textbf{Self-Supervised Methods} leverage images' semantic consistency to expand the source model's input and contrastive learning approaches to improve model generation further. ProtoTransfer \cite{Medina2020SelfSupervisedPT} uses self-supervision to train an embedding function, leading to fast adaptation to new domains. STCDFS \cite{Liu2021SelfTaughtCF} divides the domain adaptation task into inner and outer tasks. The model first self-supervises the inner task (through image rotation or background swapping) and then solves the few-shot learning problem in the outer task. TL-SS \cite{Yuan2022TaskLevelSF} follows the episodic training approach, advocating for task-level self-supervision to handle the domain discrepancy problem.

\subsubsection{Discussion and Summary}
The single-source solution proposes that a single-source domain model is sufficient for training, implying fewer requirements from different domains than multi-source approaches. However, this raises the question of whether the target domain data can be reached. If accessible, both supervised and unsupervised access are possible options. If forbidden,  self-supervision can be adopted to enrich data distribution. Single-source cross-domain few-shot learning is one of the most pervasive studies in this area, and venturing into target-domain data prohibition is the most exciting challenge here. In the following section, we will examine solutions to this issue.

\section{DIFFERENT SOLUTION OF CDFS}\label{sec:SOLUTION}

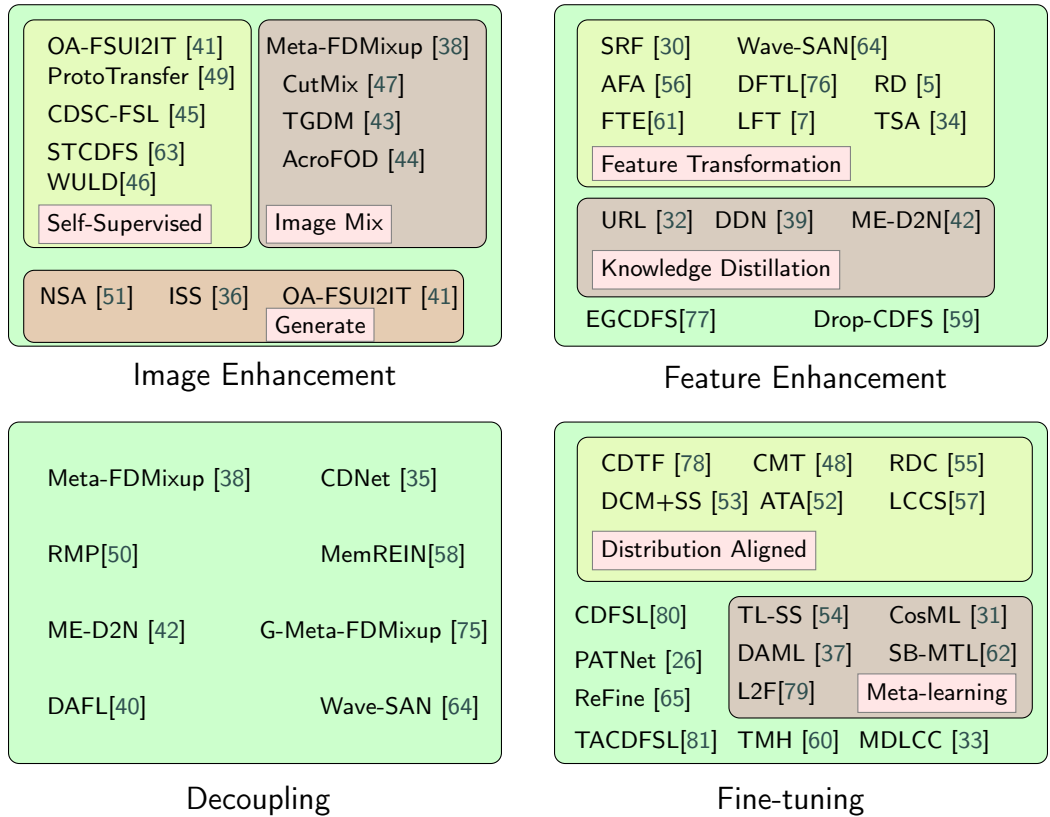
\begin{figure}[h]
		\centering
\begin{tikzpicture}

  \node[anchor=north west] at (-3,0) [rectangle, draw=black, minimum width =6.5cm,minimum height = 4.5cm,fill=green!40,rounded corners,fill opacity=0.5]{};
    
  \node[anchor=north west] at (-2.8,-0.2) [rectangle, draw=black, minimum width =3cm,minimum height = 3cm,fill=yellow!40,rounded corners,fill opacity=0.5]{};

    \node[anchor=west] at (-2.6,-2.9) [rectangle,draw,fill=pink!40, node font=\small]{\tblack{Self-Supervised}};
    
      \node[anchor=north west] at (-2.6,-0.3) [rectangle]{\tblack{OA-FSUI2IT \cite{Zhao2022OAFSUI2ITAN}
}};
      \node[anchor=north west] at (-2.6,-1.2) [rectangle]{\tblack{CDSC-FSL \cite{Chen2022CrossDomainCF}
}};
      \node[anchor=north west] at (-2.6,-2.1) [rectangle]{\tblack{WULD\cite{Yao2021CrossdomainFL}
}};
      \node[anchor=north west] at (-2.6,-1.7) [rectangle]{\tblack{STCDFS \cite{Liu2021SelfTaughtCF}
}};
      \node[anchor=north west] at (-2.6,-0.7) [rectangle]{\tblack{ProtoTransfer \cite{Medina2020SelfSupervisedPT}
}};

      \node[anchor=north west] at (0.3,-0.2) [rectangle, draw=black, minimum width =3cm,minimum height = 3cm,fill=purple!40,rounded corners,fill opacity=0.5]{};
      
          \node[anchor=west] at (0.4,-2.9) [rectangle,draw,fill=pink!40, node font=\small]{\tblack{Image Mix}};
    
      \node[anchor=north west] at (0.28,-0.3) [rectangle]{\tblack{Meta-FDMixup \cite{fu2021meta}
}};
      \node[anchor=north west] at (0.5,-0.8) [rectangle]{\tblack{CutMix \cite{Nakamura2022FewshotAO}
}};
      \node[anchor=north west] at (0.5,-1.3) [rectangle]{\tblack{TGDM \cite{Zhuo2022TGDMTG}
}};
      \node[anchor=north west] at (0.5,-1.8) [rectangle]{\tblack{AcroFOD \cite{Gao2022AcroFODAA}
}};

      \node[anchor=north west] at (-2.8,-3.5) [rectangle, draw=black, minimum width =5.8cm,minimum height = 0.95cm,fill=red!40,rounded corners,fill opacity=0.5]{};
         
      \node[anchor=north west] at (0.4,-4) [rectangle,draw,fill=pink!40, node font=\small]{\tblack{Generate}};
    
      \node[anchor=north west] at (-2.7,-3.6) [rectangle]{\tblack{NSA \cite{Liang2021BoostingTG}
}};
      \node[anchor=north west] at (-1,-3.6) [rectangle]{\tblack{ISS \cite{Xu2022CrossDomainFC}
}};

\node[anchor=north west] at (0.5,-3.6) [rectangle]{\tblack{OA-FSUI2IT \cite{Zhao2022OAFSUI2ITAN}
}};

    \node[anchor=north west] at (4.2,0) [rectangle, draw=black, minimum width =6.5cm,minimum height = 4.5cm,fill=green!40,rounded corners,fill opacity=0.5]{};

      \node[anchor=north west] at (4.5,-0.2) [rectangle, draw=black, minimum width =5.5cm,minimum height = 2.2cm,fill=yellow!40,rounded corners,fill opacity=0.5]{};

     \node[anchor=west] at (4.7,-2.1) [rectangle,draw,fill=pink!40, node font=\small]{\tblack{Feature Transformation}};
    
      \node[anchor=north west] at (4.7,-0.3) [rectangle]{\tblack{SRF \cite{Dvornik2020SelectingRF}
}};
      \node[anchor=north west] at (4.7,-0.8) [rectangle]{\tblack{AFA \cite{Hu2022AdversarialFA}
}};
      \node[anchor=north west] at (4.7,-1.3) [rectangle]{\tblack{FTE\cite{Liu2020FeatureTE}
}};
      \node[anchor=north west] at (6.5,-0.3) [rectangle]{\tblack{Wave-SAN\cite{Fu2022WaveSANWB}
}};
      \node[anchor=north west] at (6.5,-0.8) [rectangle]{\tblack{DFTL\cite{Yalan2021CrossDomainFC}
}};
      \node[anchor=north west] at (6.5,-1.3) [rectangle]{\tblack{LFT \cite{Tseng2020CrossDomainFC}
}};
      \node[anchor=north west] at (8.3,-1.3) [rectangle]{\tblack{TSA \cite{Li2022CrossdomainFL}
}};
      \node[anchor=north west] at (8.3,-0.8) [rectangle]{\tblack{RD \cite{Wang2022RememberTD}
}};


      \node[anchor=north west] at (4.5,-2.55) [rectangle, draw=black, minimum width =5.5cm,minimum height = 1.3cm,fill=purple!40,rounded corners,fill opacity=0.5]{};

     \node[anchor=west] at (4.7,-3.5) [rectangle,draw,fill=pink!40, node font=\small]{\tblack{Knowledge Distillation}};
    
      \node[anchor=north west] at (4.7,-2.6) [rectangle]{\tblack{URL \cite{Li2021UniversalRL}
}};
      \node[anchor=north west] at (6.2,-2.6) [rectangle]{\tblack{DDN \cite{Islam2021DynamicDN}
}};
      \node[anchor=north west] at (8,-2.6) [rectangle]{\tblack{ME-D2N\cite{Fu2022MED2NMD}
}};

   \node[anchor=north west] at (4.5,-3.9) [rectangle]{\tblack{EGCDFS\cite{Sun2020ExplanationGuidedTF}
}};
      \node[anchor=north west] at (7.5,-3.9) [rectangle]{\tblack{Drop-CDFS \cite{Tu2021ADS}
}};

\node[anchor=north west] at (-3,-5.5) [rectangle, draw=black, minimum width =6.5cm,minimum height = 4.5cm,fill=green!40,rounded corners,fill opacity=0.5]{};
        
      \node[anchor=north west] at (-2.6,-6) [rectangle]{\tblack{Meta-FDMixup \cite{fu2021meta}
}};

      \node[anchor=north west] at (-2.6,-7) [rectangle]{\tblack{RMP\cite{Zou2020RevisitingMP}
}};

 \node[anchor=north west] at (-2.6,-8) [rectangle]{\tblack{ME-D2N \cite{Fu2022MED2NMD}
}};
      \node[anchor=north west] at (-2.6,-9) [rectangle]{\tblack{DAFL\cite{Zhao2020DomainAdaptiveFL}
}};
      \node[anchor=north west] at (1,-6) [rectangle]{\tblack{CDNet \cite{Zou2022LearntoDecomposeCD}
}};
      \node[anchor=north west] at (1,-7) [rectangle]{\tblack{MemREIN\cite{Xu2022MemREINRT}
}};

      \node[anchor=north west] at (0.2,-8) [rectangle]{\tblack{G-Meta-FDMixup \cite{Fu2022GeneralizedMC}
}};
      \node[anchor=north west] at (1,-9) [rectangle]{\tblack{Wave-SAN \cite{Fu2022WaveSANWB}
}};

    \node[anchor=north west] at (4.2,-5.5) [rectangle, draw=black, minimum width =6.5cm,minimum height = 4.5cm,fill=green!40,rounded corners,fill opacity=0.5]{};

      \node[anchor=north west] at (4.5,-5.7) [rectangle, draw=black, minimum width =6cm,minimum height = 1.9cm,fill=yellow!40,rounded corners,fill opacity=0.5]{};

     \node[anchor=west] at (4.7,-7.2) [rectangle,draw,fill=pink!40, node font=\small]{\tblack{Distribution Aligned}};
    
      \node[anchor=north west] at (4.7,-5.8) [rectangle]{\tblack{CDTF \cite{Lu2022CrossdomainFS}
}};
      \node[anchor=north west] at (6.7,-5.8) [rectangle]{\tblack{CMT \cite{Teshima2020FewshotDA}
}};
      \node[anchor=north west] at (8.5,-5.8) [rectangle]{\tblack{RDC \cite{Li2021RankingDC}
}};
      \node[anchor=north west] at (4.7,-6.3) [rectangle]{\tblack{DCM+SS \cite{Tao2022PoweringFI}
}};

      \node[anchor=north west] at (6.8,-6.3) [rectangle]{\tblack{ATA\cite{Wang2021CrossDomainFC}
}};
      \node[anchor=north west] at (8.5,-6.3) [rectangle]{\tblack{LCCS\cite{Zhang2022FewShotAO}
}};

      \node[anchor=north west] at (6.5,-7.8) [rectangle, draw=black, minimum width =4cm,minimum height = 1.6cm,fill=purple!40,rounded corners,fill opacity=0.5]{};

     \node[anchor=west] at (8.2,-9.08) [rectangle,draw,fill=pink!40, node font=\small]{\tblack{Meta-learning}};
    
      \node[anchor=north west] at (6.5,-7.8) [rectangle]{\tblack{TL-SS \cite{Yuan2022TaskLevelSF}
}};
      \node[anchor=north west] at (6.5,-8.3) [rectangle]{\tblack{DAML \cite{Lee2022DomainAgnosticMF}
}};
      \node[anchor=north west] at (6.5,-8.8) [rectangle]{\tblack{L2F\cite{Baik2021LearningTF}
}};

      \node[anchor=north west] at (8.5,-7.8) [rectangle]{\tblack{CosML \cite{Peng2020CombiningDM}
}};

      \node[anchor=north west] at (8.5,-8.3) [rectangle]{\tblack{SB-MTL\cite{Cai2020SBMTLSM}
}};

   \node[anchor=north west] at (4.35,-7.8) [rectangle]{\tblack{CDFSL\cite{Lin2021ModularAF}
}};
   \node[anchor=north west] at (4.35,-8.4) [rectangle]{\tblack{PATNet \cite{Lei2022CrossDomainFS}
}};
   \node[anchor=north west] at (4.35,-9.45) [rectangle]{\tblack{TACDFSL\cite{Zhang2022TACDFSLTA}
}};
      \node[anchor=north west] at (4.35,-8.9) [rectangle]{\tblack{ReFine \cite{Oh2022ReFineRB}
}};
      \node[anchor=north west] at (6.5,-9.45) [rectangle]{\tblack{TMH \cite{Jiang2020ATM}
}};
      \node[anchor=north west] at (8.1,-9.45) [rectangle]{\tblack{MDLCC \cite{Xiao2020MultiDomainLF}
}};

\node[anchor=west] at (-1.5,-4.9) [rectangle, node font=\large]{\tblack{Image Enhancement}};

\node[anchor=west] at (5.5,-4.9) [rectangle, node font=\large]{\tblack{Feature Enhancement}};

\node[anchor=west] at (-0.8,-10.5) [rectangle, node font=\large]{\tblack{Decoupling}};

\node[anchor=west] at (6.2,-10.5) [rectangle, node font=\large]{\tblack{Fine-tuning}};

\end{tikzpicture}
\caption{Different solution of CDFS. Image Enhancement directly mix the image and generate the new image.
      Feature Enhancement use feature-wise transformation and knowledge distillation to achieve enhancement.
      Finetuning focus on the adaptability feature training.
      Decoupling solutions expect to get the domain-irrelevant and domain-specific feature.}
  
		\vspace{-0.3cm}
		\label{fig:diffsolu}
	\end{figure}

We divide the solution of CDFS into four categories: Feature-Enhancement, Image-Enhancement, Decompose, and Finetuning (see Fig. \ref{fig:diffsolu}). The Feature-Enhancement Based methods seek to improve the model's performance through expanded feature diversity. In contrast, the Image-Enhancement Based methods leverage the usual mixing or pasting techniques to generate additional images. Decompose Based methods consider both domain-specific and domain-agnostic features, and finally, Finetuning Based methods strive to obtain highly adaptable features to the new domain. We further discuss these solutions, each in its dedicated section (\ref{Feature-Enhancement}, \ref{Image-Enhancement}, \ref{Decompose}, and \ref{Finetuning}), below.
\subsection{Feature-Enhancement Based} \label{Feature-Enhancement}
\begin{figure}[h]
		\centering
\includegraphics[width=0.9\textwidth,trim={0cm 0cm 0cm 0cm}, clip]{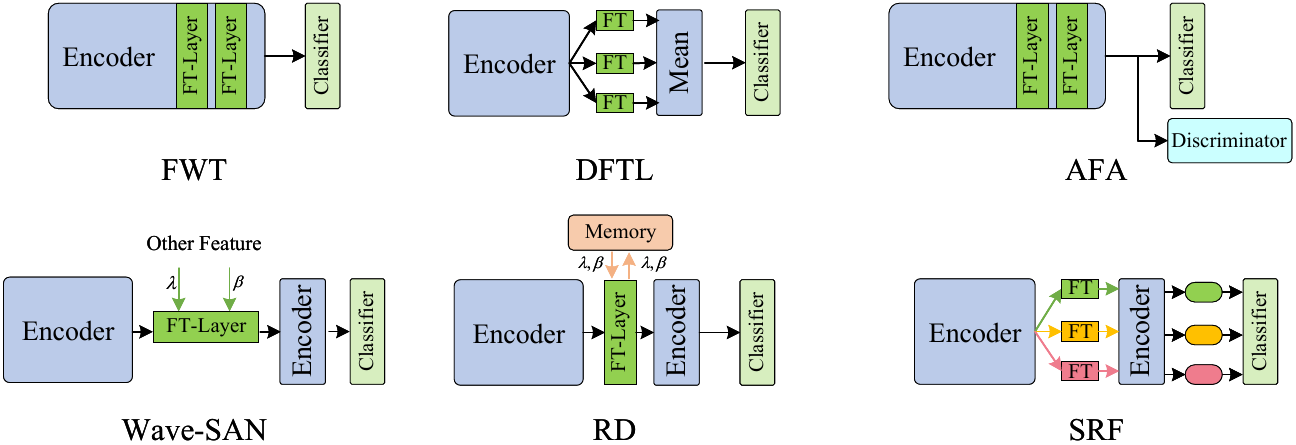}
    		\caption{The different feature transformation in the CDFS.
      FWT \cite{Tseng2020CrossDomainFC},  DFTL\cite{Yalan2021CrossDomainFC}, AFA \cite{Hu2022AdversarialFA} and SRF \cite{Dvornik2020SelectingRF} use the learnable $\lambda$ and $\beta$ to achieve knowledge transfer, and each domain data have a FT-layer in the SRF \cite{Dvornik2020SelectingRF}.
      Wave-SAN \cite{Fu2022WaveSANWB} and RD \cite{Wang2022RememberTD} mix the feature distribution to restructure the feature.
		}
		\vspace{-0.3cm}
		\label{fig:diff_fwt}
  \vspace{-0.3cm}
	\end{figure}

After training in the base classes, the few-shot model must learn the novel classes in a new domain with limited annotation. In such cases, the feature quality directly influences the transfer effect. To address this, the CDFS researchers adopt feature transformation, as illustrated in Fig. \ref{fig:diff_fwt}, or knowledge distillation to enhance the feature quality.


\textbf{Feature Transformation Enhancement} mainly changes the feature distribution, which can make the feature easily transfer \cite{Fu2022WaveSANWB,Liu2020FeatureTE}.
The normalized strategy adds noise or prior distribution into the original feature. The training process contains normalization and reconstructing features using Eq. \ref{eq:feature_transform}:
\begin{equation}
    \begin{split}
         & f_b^{norm}=\dfrac{f_b-\mu_b}{v_b} \\
         & f_b^{reco}=\lambda f_b^{norm} + \beta
    \end{split}
    \label{eq:feature_transform}
\end{equation}

where $f_b$,$\mu_b$, and $v_b$ are backbone features, channel means and channel variance. After the feature normalization, use $\lambda$ and $\beta$ to reconstruct the backbone feature.
If $\lambda$ and $\beta$ in Eq. \ref{eq:feature_transform} from learnable parameters, the feature-wise transformer focuses on the feature distribution improvement.
FWT \cite{Tseng2020CrossDomainFC} insert a feature-wise transformation layer after the batch normalization layer in the feature encoder, which can effectively augment the intermediate feature. FWT \cite{Tseng2020CrossDomainFC} will remove the feature-wise transformation layer after the training.
And DFTL\cite{Yalan2021CrossDomainFC} propose a diversified feature transformation that parallels several feature transformers.  The final layer averages the  FT-layer output and gets the final prediction.
The single-source network also can use feature transformation to simulate unseen domain distribution.
In the AFA \cite{Hu2022AdversarialFA} network, the domain discriminator is learned by recognizing the augmented features (unseen domain) from the original ones (seen domain).
In summary, FWT \cite{Tseng2020CrossDomainFC},  DFTL\cite{Yalan2021CrossDomainFC} and AFA \cite{Hu2022AdversarialFA} use the learnable $\lambda$ and $\beta$ to achieve knowledge transfer.

The $\lambda$ and $\beta$  could come from other instance's normalized features.
Wave-SAN \cite{Fu2022WaveSANWB} augment source images by swapping the styles of their low-frequency components with each other.
Wave-SAN \cite{Fu2022WaveSANWB} follows the replacing strategy and completely restructures the original feature with another instance feature's distribution.
This may lead to fluctuation, especially if the datasets have a domain shift. 
The mixing strategy can serve as an improved version compared, where the network uses the mixed tensor to restructure the feature, as the Eq. \ref{equ:mix}.
\begin{equation}
    \begin{split}
       &\lambda^{mix}=\alpha \mu_b + (1-\alpha)\mu_o,
	 \beta^{mix}=\alpha v_b + (1-\alpha)v_o, \\
	& f_{b}^{reco}=\lambda^{mix}f_{b}^{norm} + \beta^{mix}
      \end{split}
  	\label{equ:mix}
\end{equation}
where $\alpha$ controls the retention ratio of original features, $\mu_o$ and $v_o$ are channel mean and variance from other image.
Specifically, RD \cite{Wang2022RememberTD} uses a Memory-Bank to mix the instance feature, which means the $\mu_o$ and $v_o$ in Eq. \ref{equ:mix} from the Memory-Bank which collect different source stylization information, the Memory-Bank can load the source knowledge into the target domain which can directly relive the domain shift.
For the multiple-source network, SRF \cite{Dvornik2020SelectingRF} append different FT-layer for each domain.
Using a parametric network family to obtain multi-domain representations.

\textbf{Knowledge Distillation Enhancement} train the teacher and student networks. The student network's learning target is to make the feature more robust than the teacher network.
URL \cite{Li2021UniversalRL} train one teacher network for each domain data, then use a student network to distill the domain knowledge from each teacher network.
DDN \cite{Islam2021DynamicDN} learns a teacher network with weak augmented target data, and the target domain's knowledge will be distilled from the teacher to the final transfer student network.
The distillation process also can load target domain distribution into the source training process. 
Similarly, ME-D2N \cite{Fu2022MED2NMD} also uses Knowledge Distillation to decompose the source and target domain, which can effectively improve the CDFS accuracy.

\subsection{Image-Enhancement Based} \label{Image-Enhancement}
\begin{figure}[h]
		\centering
\includegraphics[width=0.7\textwidth,trim={0cm 0cm 0cm 0cm}, clip]{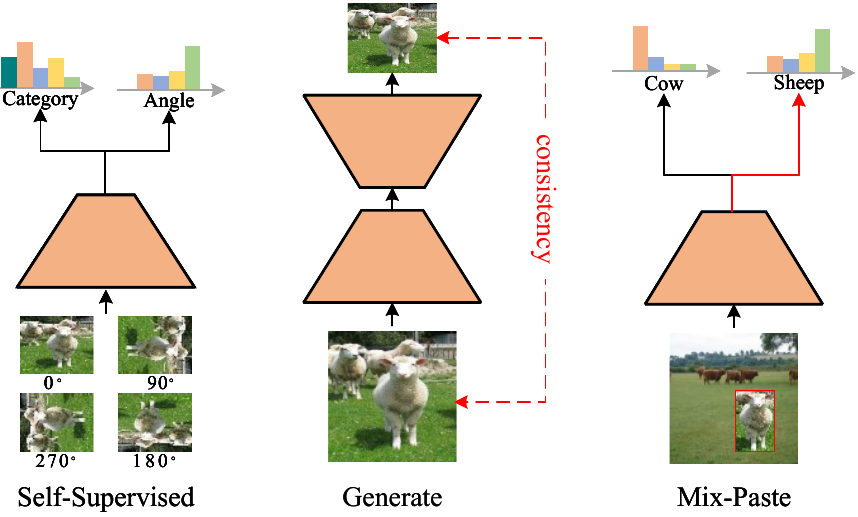}
    		\caption{Image Enhancement Based.
      Self-supervised generate the new labels for model training, and generate methods use the consistency loss to constrain the model, Mix-Paste methods mix the different image which can effectively improve the data diversity.
		}
		\vspace{-0.3cm}
		\label{fig:image_enh}
	\end{figure}

Image enhancement-based methods directly enhance the image data, as shown in Fig. \ref{fig:image_enh}. Compared with feature enhancement-based methods, it is simpler and changes the image directly. The self-supervised method transforms the original image to generate new labels for model training, providing another perspective for model observation data. At the same time, Mix-Paste methods mix the different images to enrich data distribution. The generation strategy is another enhancement strategy, which generates new images through the encoder-decoder and inputs them into the network to improve the data diversity.

\textbf{Self-Supervised Image-Enhancement} generates the new instance through images' semantic consistency. It is practical to improve data quality because this process is label-free \cite{Medina2020SelfSupervisedPT}.
OA-FSUI2IT \cite{Zhao2022OAFSUI2ITAN} assume that the image semantic is translation irrelevant. Applying flipping, translation, and linear scaling on the image in the training process will not change the image content. The deep model should obtain a similar prediction.
WULD \cite{Yao2021CrossdomainFL} train the few-shot model with the labeled and unlabeled data.
The model must recognize the labeled classes and predict the rotated degrees in the labeled and unlabeled data.
As the auxiliary task, image rotation can improve the effectiveness of the main task.
Similar to WULD \cite{Yao2021CrossdomainFL},  STCDFS \cite{Liu2021SelfTaughtCF} also uses rotation to improve the few-shot model effect.
It designs the inner task in the few-shot support set. The model first predicted the image rotation degrees and labeled classes. The model gets the class prototypes with the original support image and finally uses the prototypes to recognize the query set.
In the source unlabeled CDFS, ProtoTransfer \cite{Medina2020SelfSupervisedPT} first transforms the unlabeled batch. Each sample will generate $|Q|$ transform samples. The transformed feature needs to be concentrated around the original sample. Other original samples in the batch are regarded as negative, and the original sample is regarded as positive. Finally, in the target domain, the extracted image features will serve as the classifier weight for fine-tuning.


\textbf{Mix-Paste Image-Enhancement} \cite{Nakamura2022FewshotAO} is the most simple methods which cut few target data and past into the source data, this help the model learn the target domain knowledge.
\begin{figure}[h]
		\centering
\includegraphics[width=0.9\textwidth,trim={0cm 0cm 0cm 0cm}, clip]{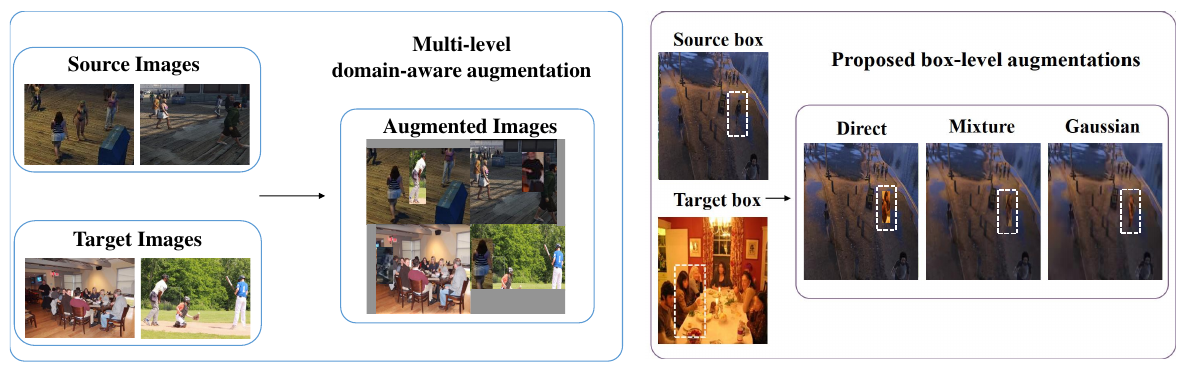}
    		\caption{The different mix methods proposed by AcroFOD \cite{Gao2022AcroFODAA}.
		}
		\vspace{-0.3cm}
		\label{fig:AcroFOD_mix}
	\end{figure}
And AcroFOD \cite{Gao2022AcroFODAA} propose two different mix methods to improve the effect of few-shot object detection as shown in Fig. \ref{fig:AcroFOD_mix}.
The left Image-level Domain-aware Augmentation generates the new image with several domain instances. And the right Box-level Domain-aware Augmentation first cuts the foreground part of the instance, then pastes the corresponding box into another instance.
Moreover, Meta-FDMixup \cite{fu2021meta} uses the super-parameter $\lambda$ to control the source and target ratio in the mixed image.
After mixing, there will be source, target, and mixed three types of images.
Due to the target labels being accessible, the source-mix pairs and target-mix pairs will run the few-shot classification and calculate the loss. The source and target mixing can effectively improve the data distribution and get more target data to fit the model.
Compare the stiffly paste methods, TGDM \cite{Zhuo2022TGDMTG} aim to learn a dynamic mix ratio via input the validation loss to a generation network, in other words, the $\lambda$ in the Meta-FDMixup \cite{fu2021meta} is learnable.

\textbf{Generate Image-Enhancement} generate a new image with the Encoder-Decoder method.
NSAE \cite{Liang2021BoostingTG}  propose to take reconstructed images from auto-encoder as noisy inputs and let the model further predict
their labels. In the target testing stage, first performs reconstruction task on the novel dataset and then the encoder is fine-tuned for classification.
Compare with 
NSAE \cite{Liang2021BoostingTG} vallina encoder-decoder,
ISS \cite{Xu2022CrossDomainFC} combines the feature transformation and generates it to solve the CDFS problem. The labeled and unlabelled domain features are mixed in the training stage. Then the mixed features are decoded to generate images, which are constrained by content perception loss and style loss to ensure that the semantic information is consistent with labeling, and the style is consistent with unlabelled. Finally, the enhanced dataset is used for few-shot training and transferred to the new domain.
Meanwhile, OA-FSUI2IT \cite{Zhao2022OAFSUI2ITAN} focuses on object detection and proposes an image-translation module that generates the source content consistency and target style consistency image.
ISS \cite{Xu2022CrossDomainFC} and OA-FSUI2IT \cite{Zhao2022OAFSUI2ITAN} have similar inspiration.
Specifically,
Perceptual loss measures the perceptual similarity of two samples, shown in the Eq. \ref{eq:perceptual}.
\begin{equation}
    \begin{split}
         \mathcal{L}_{percep}=\sum_{l \in \mathcal{S}} ||f_l(x) - f_l(\hat{x})||_{2}^{2}
    \end{split}
    \label{eq:perceptual}
\end{equation}
where $f_l$ denotes feature from the converged network $\mathcal{S}$, $\mathcal{L}_{percep}$ require the input $x$ and generated image $\hat{x}$ share the same content or semantic.
Style loss measures
the differences between covariances of the feature maps, which can reduce the stylisation from different images, as shown in Eq. \ref{eq:style}.
\begin{equation}
    \begin{split}
         \mathcal{L}_{style}=||G_{j}(x)-G_{j}(\hat{x})||_F^{2}
    \end{split}
    \label{eq:style}
\end{equation}
where $G_{j}$ get the gram matrix of $j$-th layer.
And OA-FSUI2IT \cite{Zhao2022OAFSUI2ITAN} additional use adversarial networks to improve the generation effect.
Generate Image-Enhancement is the specific self-supervised version, but it generate new image compare the single image rotation and mask.

\subsection{Decompose Based} \label{Decompose}

\begin{figure}[h]
		\centering
		\includegraphics[height=0.25\textwidth,trim={0cm 0cm 0cm 0cm}, clip]{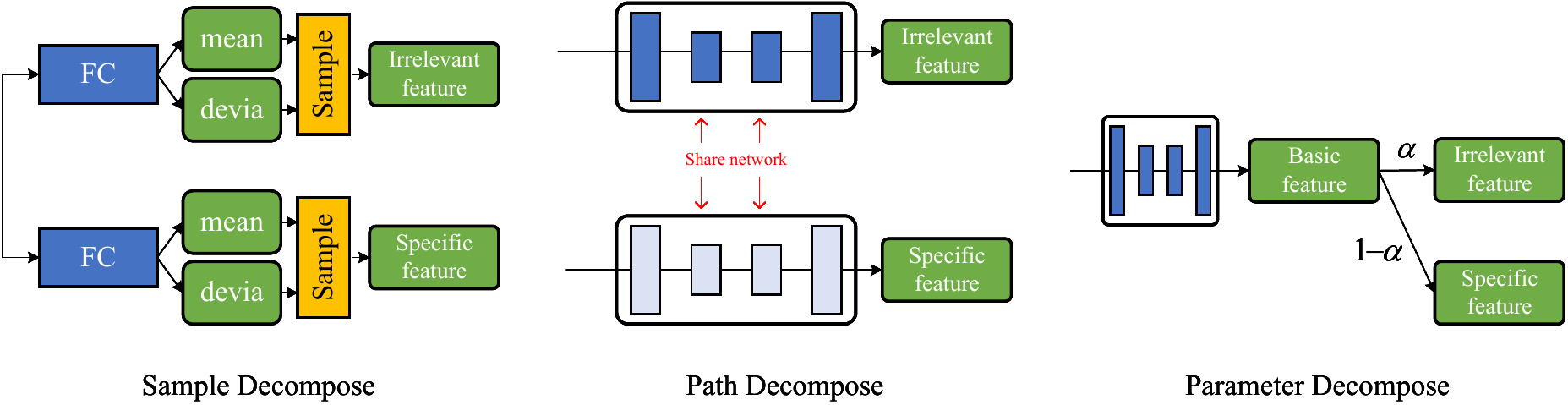}
    		\caption{Different decompose methods.
      Sample based methods extract different feature mean and variance, then sample the specific feature.
      Path based methods design several forward path for feature passing, and each path decompose different feature.
      Parameter based methods learn the decompose parameter, e.g.,  $\alpha$  and use the it to filter the feature.
		}
		\vspace{-0.3cm}
		\label{fig:diff_decompose}
	\end{figure}

Methods think different features should be decomposed, as shown in Fig. \ref{fig:diff_decompose}. 
Due to the data of CDFS from a different domain, the mixed feature space disturbs the model effect. To decompose these features, researchers \cite{Zou2022LearntoDecomposeCD,Zhao2020DomainAdaptiveFL,Xu2022MemREINRT} except to get different feature expressions. 
E.g.,  The domain-specific and domain-irrelevant features, The irrelevant feature mainly collects the domain-agnostic discriminate information, which means the feature has excellent transfer ability across the domain. As a complement, the irrelevant feature can provide the related domain information. The specific and irrelevant features adapt to each other through adversarial learning.

Meta-FDMixup serious \cite{fu2021meta,Fu2022GeneralizedMC} design a disentangle module which generates the mean and variance of irrelevant and specific features as shown in the Fig. \ref{fig:diff_decompose} left part, then use the corresponding mean and variance to sample the domain feature.
The disentangle module is inspired by VAE (Variation Auto-Encoder), and lies in the best learning to disentangle the domain-specific
and domain-irrelevant features, thus alleviating the domain shift problem in CDFS.

Different from the Sample base methods, as shown in the Fig. \ref{fig:diff_decompose} middle part, Path-based use several forward paths to learn the different feature expressions, such as ME-D2N \cite{Fu2022MED2NMD} extra learn domain-specific gate forward path which can assign each filter to only one specific domain in a learnable way.
The structure of Wave-SAN \cite{Fu2022WaveSANWB} contains stand and style-augmented forward. And  Wave-SAN \cite{Fu2022WaveSANWB} aims to enable the decomposition of visual representations into low-frequency components, such as shape and style, and high-frequency components, e.g., texture. The various feature produced required to share the same semantics can effectively improve the feature quality of CDFS.

The Parameter methods use the learnable tensor to filter screen the feature, as shown in the right part of Fig. \ref{fig:diff_decompose}.
CDNet \cite{Zou2022LearntoDecomposeCD} propose a serial framework. Each step uses the Learn-to-Decompose (LD) Module to filter the domain feature and complete the feature decomposition.
A MemREIN \cite{Xu2022MemREINRT} focuses on the channel decomposition, the discriminative feature will be registered into the memory bank.

Besides these domain decomposing methods, researchers also study the expression at the class level.
RMP \cite{Zou2020RevisitingMP} considers that the mid-level feature is more meaningful for the transferring between distant domains. Class mid-level features must predict the class discriminates features to supply the mid-level features. As the middle feature lacks the discrimination, RMP \cite{Zou2020RevisitingMP} first extract the class feature, then use the features of other classes to reconstruct the current class, and finally use the existing features to subtract the reconstructed features in the cosine space to obtain the class discriminate features.


\subsection{Fine-tuning Based} \label{Finetuning}
Fine-tuning Based methods focus on the feature transfer \cite{Xiao2020MultiDomainLF} and parameters adaptability \cite{Oh2022ReFineRB}.
Fine-tuning aims to get a robust feature with high transfer ability \cite{Zhang2022TACDFSLTA}, expect to transform domain-specific features into irrelevant metric spaces, which will reduce the adverse effects of domain shift. Since the metric space is invariant, it is easier for downstream segmentation modules to predict in such a stable space.
We group Fine-tuning Based methods into \textbf{Meta-Learning Methods}, \textbf{Distribution Aligned Methods}.
\subsubsection{Meta-Learning Fine-tuning}
\begin{figure}[h]
		\centering
	\includegraphics[height=0.2\textwidth,trim={0cm 0cm 0cm 0cm}, clip]{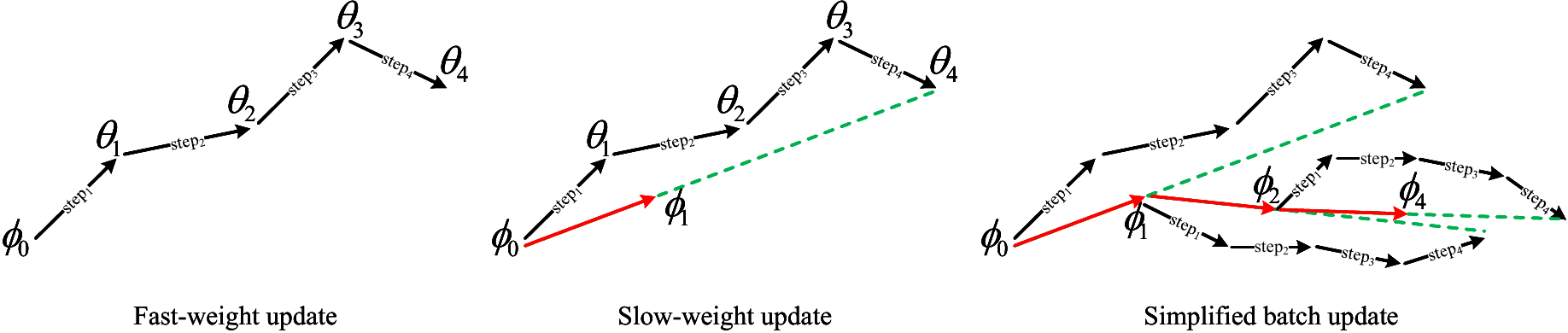}
    		\caption{The parameter updating strategy of the meta-generation network, where $\theta$ is the fast-weight and $\phi$ is the slow-weight.
		}
		\label{fig:DCM+SS}
	\end{figure}
The core idea of Meta-Learning is 'learn to learn,' which means the optimized target is how to collect the experience faced by the new task.
Compared with traditional training methods, the meta-learning process has fast-weights and slow-weights. The fast-weight solves the specific task problem. The slow-weight needs to collect experience, as shown in Fig. \ref{fig:DCM+SS}.

The key design of meta-learning is task sampling. DAML \cite{Lee2022DomainAgnosticMF} sample the tasks from multiple domains,  the model learns domain-agnostic initial parameters, which would adapt to novel classes in unseen domains during meta-testing.
And CosML \cite{Peng2020CombiningDM} directly combines multiple domain parameters as the transfer initialization, its training process on each separate domain compared with DAML \cite{Lee2022DomainAgnosticMF}.
And except for the slow and fast optimized weight, TL-SS \cite{Yuan2022TaskLevelSF} uses a Weight Generator to predict the parameters of a high-level network, which require the model to generate proper parameters and enables the encoder to flexibly to any unseen tasks. 
For parameters initialization in the meta-learning, forcibly
sharing an initialization can lead to conflicts among tasks and the compromised (undesired by tasks) location on the optimization landscape, thereby hindering task adaptation.
L2F \cite{Baik2021LearningTF} propose task-and-layer-wise attenuation on the compromised initialization to reduce its
adverse influence on task adaptation.

\subsubsection{Distribution Aligned Fine-tuning}
\begin{figure}[h]
		\centering
	\includegraphics[height=0.3\textwidth,trim={0cm 0cm 0cm 0cm}, clip]{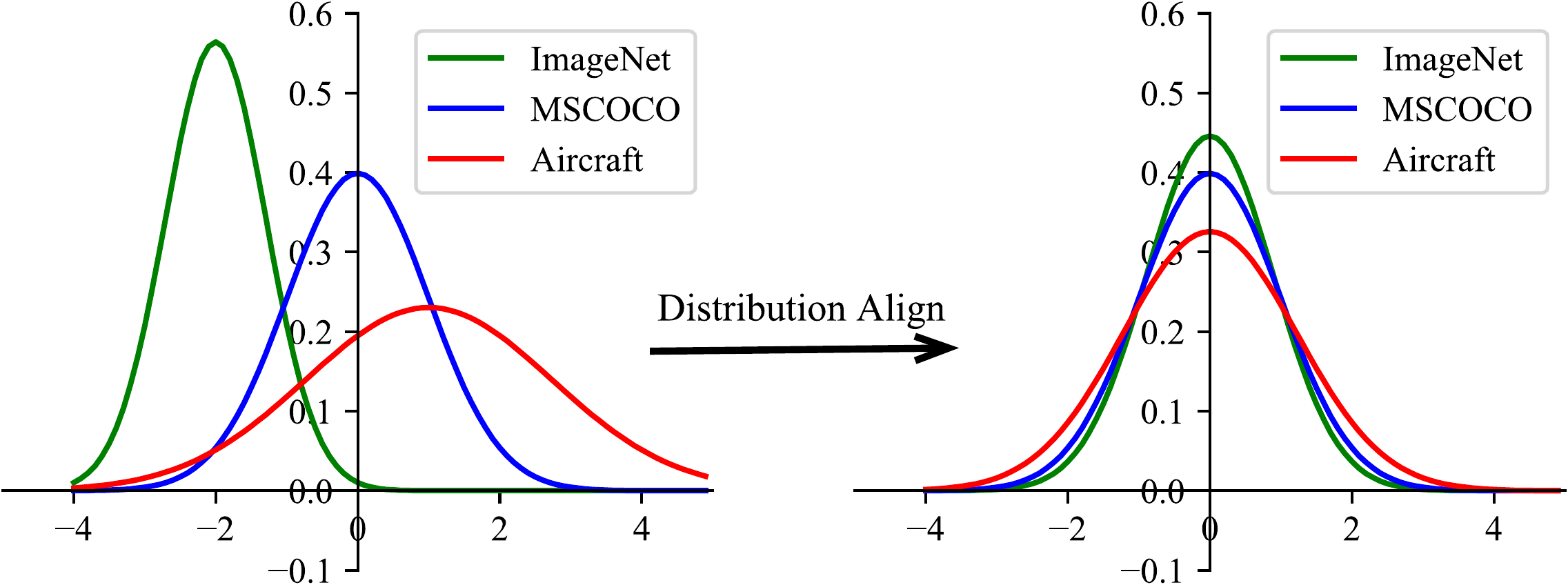}
    		\caption{Different scenario have disunion distribution, which may reflect in the input image or feature level(left part).
      After Distribution Align, optimize the model could in a unified space, and have better transfer ability(right part).
		}
		\label{fig:DCM+SS}
	\end{figure}
The distribution between the source and target domains in the CDFS is different, as shown in Fig. \ref{fig:DCM+SS}. To address the influence of this difference, CMT \cite{Teshima2020FewshotDA} proposed mechanism transfer, a meta-distributional scenario in which the data-generating mechanism is invariant across domains. CMT \cite{Teshima2020FewshotDA} enabled domain adaptation among disparate distributions without relying on parametric assumptions. To prevent the pre-trained representation from being biased towards the source domain, RDC \cite{Li2021RankingDC} constructed a non-linear subspace to minimize task-irrelevant features while also retaining more transferrable discriminative information through a hyperbolic tangent transformation. In other words, the source data was aligned between the original space and the non-linear subspace, resulting in more transferable discriminant information. CDTF \cite{Lu2022CrossdomainFS} focused on aligning the support and query sets and proposed to transductive fine-tune the base model on a set of query images under the few-shot setting. The core idea was to implicitly guide query image segmentation using support labels. 

As the base unit of meta-learning, the task distribution directly influences the model's effectiveness. ATA \cite{Wang2021CrossDomainFC} sought to improve the robustness of the inductive bias through task augmentation. The network adaptively generates 'challenging' tasks using different inductive biases. Task-level augmentation can increase the diversity of training tasks and improve the model's robustness under domain shift. At the feature level, researchers \cite{Zhang2022FewShotAO} attempted to transfer the source distributions into the target domain.

\clearpage
\section{BENCHMARK}
\label{BENCHMARK}

\subsection{CDFS Classification.} 
BSCD-FSL \cite{Guo2019ABS} propose the Broader Study of Cross-Domain Few-Shot Learning (BSCD-FSL)
benchmark, consisting of image data from diverse image acquisition methods. 
There are mini-ImageNet \cite{Ravi2016OptimizationAA}, CropDiease \cite{Mohanty2016UsingDL}, EuroSAT \cite{Helber2019EuroSATAN}, ISIC \cite{Tschandl2018TheHD}, ChestX \cite{Wang2017ChestXRay8HC} datasets in BSCD-FSL benchmark. Since the mini-ImageNet \cite{Ravi2016OptimizationAA} has 100 classes where each class has 600 images, BSCD-FSL selects it as the source domain and transfers the model knowledge to the rest domains.

Meanwhile, LFT \cite{Tseng2020CrossDomainFC} proposes a CDFS benchmark that contains five datasets. LFT design the leave-one-out experiments setting, which selects one domain from the CUB \cite{Wah2011TheCB}, Cars \cite{Krause20133DOR}, Places \cite{Zhou2018PlacesA1}, and Plantae \cite{Horn2017TheIS} as the unseen domain for the evaluation, the mini-ImageNet \cite{Ravi2016OptimizationAA} and the remaining domains serve as the seen domains for training the model.

For cross-domain few-shot datasets, Meta-Dataset \cite{Triantafillou2019MetaDatasetAD} proposes a new benchmark for training and evaluating large-scale models, consists of diverse datasets and presents more realistic tasks.
Meta-Dataset is divided by domain. That is, there is no cross-category between datasets. We show the benchmark used for all the CDFS classification methods in Tab. \ref{tab:1_shot_cls} and Tab. \ref{tab:k_shot_cls}.
The most frequent experiment is the 5-way 1-shot setting. We summary the 5-way 1-shot transfer effect in the Tab. \ref{tab:1_shot_cls} and the 5-way k-shot transfer effect in the Tab. \ref{tab:k_shot_cls}.

\begin{table}[h]
\centering
 \footnotesize
\tabcolsep=5pt
\caption{The CDFS classification result of 5-way 1-shot. $^\dag$ denotes the leave-one-out experiments, $^*$ denotes the target data is accessible, $^\mathrm{T}$ denotes the transductive methods.}
\begin{tabular}{l|llllllll} 
\toprule
5-way 1-shot      & CUB         & Cars         & Places       & Plantae      & CropDiseases & EuroSAT     & ISIC       & ChestX      \\ 
\hline
MatchingNet\cite{Vinyals2016MatchingNF}$_{(\mathrm{NIPS16)}}$       & 35.89$_{\pm \mathrm{0.5}}$   & 30.77$_{\pm \mathrm{0.5}}$    & 49.86$_{\pm \mathrm{0.8}}$    & 32.70$_{\pm \mathrm{0.6}}$    & -            & -           & -          & -           \\
MN+FT\cite{Tseng2020CrossDomainFC}$_{(\mathrm{ICLR20)}}$   & 36.61$_{\pm \mathrm{0.5}}$   & 29.82$_{\pm \mathrm{0.4}}$    & 51.07$_{\pm \mathrm{0.7}}$    & 34.48$_{\pm \mathrm{0.5}}$    & -            & -           & -          & -           \\
RelationNet\cite{Sung2017LearningTC}$_{(\mathrm{CVPR18)}}$       & 42.44$_{\pm \mathrm{0.8}}$   & 29.11$_{\pm \mathrm{0.6}}$    & 48.64$_{\pm \mathrm{0.9}}$    & 33.17$_{\pm \mathrm{0.6}}$    & -            & -           & -          & -           \\
RN+FT\cite{Tseng2020CrossDomainFC}$_{(\mathrm{ICLR20)}}$   & 44.07$_{\pm \mathrm{0.8}}$   & 28.63$_{\pm \mathrm{0.6}}$    & 50.68$_{\pm \mathrm{0.8}}$    & 33.14$_{\pm \mathrm{0.6}}$    & -            & -           & -          & -           \\
GNN\cite{Satorras2017FewShotLW}$_{(\mathrm{ICLR18)}}$               & 45.69$_{\pm \mathrm{0.7}}$   & 31.79$_{\pm \mathrm{0.5}}$    & 53.10$_{\pm \mathrm{0.8}}$    & 35.60$_{\pm \mathrm{0.6}}$    & -            & -           & -          & -           \\
GNN+FT\cite{Tseng2020CrossDomainFC}$_{(\mathrm{ICLR20)}}$           & 47.47$_{\pm \mathrm{0.8}}$   & 31.61$_{\pm \mathrm{0.5}}$    & 55.77$_{\pm \mathrm{0.8}}$    & 35.95$_{\pm \mathrm{0.6}}$    & -            & -           & -          & -           \\
RN+LRP\cite{Sun2020ExplanationGuidedTF}$_{(\mathrm{ICPR20)}}$            & 42.44$_{\pm \mathrm{0.4}}$   & 29.65$_{\pm \mathrm{0.3}}$    & 50.59$_{\pm \mathrm{0.5}}$    & 34.80$_{\pm \mathrm{0.4}}$    & -            & -           & -          & -           \\
RN+LRP$^\mathrm{T}$\cite{Sun2020ExplanationGuidedTF}$_{(\mathrm{ICPR20)}}$          & 42.88$_{\pm \mathrm{0.5}}$   & 29.61$_{\pm \mathrm{0.4}}$    & 53.07$_{\pm \mathrm{0.6}}$   & 34.54$_{\pm \mathrm{0.4}}$    & -            & -           & -          & -           \\
CAN+LRP\cite{Sun2020ExplanationGuidedTF}$_{(\mathrm{ICPR20)}}$           & 46.23$_{\pm \mathrm{0.4}}$   & 32.66$_{\pm \mathrm{0.5}}$    & 56.96$_{\pm \mathrm{0.5}}$    & 38.23$_{\pm \mathrm{0.5}}$    & -            & -           & -          & -           \\
CAN+LRP$^\mathrm{T}$\cite{Sun2020ExplanationGuidedTF}$_{(\mathrm{ICPR20)}}$         & 48.35$_{\pm \mathrm{0.5}}$   & 32.35$_{\pm \mathrm{0.4}}$    & 61.60$_{\pm \mathrm{0.6}}$    & 38.48$_{\pm \mathrm{0.4}}$    & -            & -           & -          & -           \\
GNN+LRP\cite{Sun2020ExplanationGuidedTF}$_{(\mathrm{ICPR20)}}$           & 48.29$_{\pm \mathrm{0.5}}$   & 32.78$_{\pm \mathrm{0.4}}$    & 54.83$_{\pm \mathrm{0.6}}$    & 37.49$_{\pm \mathrm{0.4}}$    & -            & -           & -          & -           \\
GNN+FT$^\dag$\cite{Tseng2020CrossDomainFC}$_{(\mathrm{ICLR20)}}$       & 51.51$_{\pm \mathrm{0.8}}$ & 34.12 $_{\pm \mathrm{ 0.6}}$  & 56.31 $_{\pm \mathrm{0.8}}$  & 42.09 $_{\pm \mathrm{ 0.7}}$  & -            & -           & -          & -           \\
CosML$^\dag$\cite{Peng2020CombiningDM}$_{(\mathrm{Arxiv)}}$        & 46.89$_{\pm \mathrm{0.5}}$   & 47.74$_{\pm \mathrm{0.6}}$  & 53.96$_{\pm \mathrm{0.6}}$  & 30.93$_{\pm \mathrm{0.5}}$  & -            & -           & -          & -           \\
LRFG$^\dag$\cite{Chen2022CrossDomainFC}$_{(\mathrm{KBS22)}}$             & 52.04 $_{\pm \mathrm{ 0.7}}$ & 34.84 $_{\pm \mathrm{0.6}}$  & 57.57 $_{\pm \mathrm{ 0.8}}$  & 42.05 $_{\pm \mathrm{ 0.7}}$ & -            & -           & -          & -           \\
DFTL$^\dag$\cite{Yalan2021CrossDomainFC}$_{(\mathrm{ICAICA21)}}$             & 46.15$_{\pm \mathrm{0.7}}$   & 33.54 $_{\pm \mathrm{0.6}}$   & 51.81$_{\pm \mathrm{ 0.7}}$   & 39.97$_{\pm \mathrm{0.6}}$    & -            & -           & -          & -           \\
GNN+MR$^\dag$\cite{Xu2022MemREINRT}$_{(\mathrm{IJCAI22)}}$   & 54.26 $_{\pm \mathrm{ 0.6}}$ & 37.55 $_{\pm \mathrm{0.5}}$ & 59.98 $_{\pm \mathrm{ 0.6}}$ & 45.69 $_{\pm \mathrm{ 0.6}}$  & -            & -           & -          & -           \\
MN+MR$^\dag$\cite{Xu2022MemREINRT}$_{(\mathrm{IJCAI22)}}$ & 46.37 $_{\pm \mathrm{ 0.5}}$ & 35.65 $_{\pm \mathrm{0.5}}$  & 54.92 $_{\pm \mathrm{ 0.6}}$  & 38.82 $_{\pm \mathrm{ 0.5}}$ & -            & -           & -          & -           \\
RN+MR$^\dag$\cite{Xu2022MemREINRT}$_{(\mathrm{IJCAI22)}}$ & 52.02 $_{\pm \mathrm{ 0.5}}$ & 36.38 $_{\pm \mathrm{0.4}}$ & 54.82 $_{\pm \mathrm{ 0.6}}$  & 36.74 $_{\pm \mathrm{ 0.5}}$  & -            & -           & -          & -           \\
RN+ST\cite{Liu2021SelfTaughtCF}$_{(\mathrm{Arxiv)}}$             & 43.10 $_{\pm \mathrm{ 0.4}}$ & 32.34 $_{\pm \mathrm{0.3}}$  & 50.53 $_{\pm \mathrm{ 0.5}}$  & 33.19 $_{\pm \mathrm{ 0.3}}$  & 63.29$_{\pm \mathrm{ 0.4}}$  & 57.36$_{\pm \mathrm{0.30}}$ & 32.09$_{\pm \mathrm{0.3}}$  & 22.28$_{\pm \mathrm{0.2}}$  \\
ReFine\cite{Oh2022ReFineRB}$_{(\mathrm{CIKM22)}}$            & -           & -            & -            & -            & 68.93$_{\pm \mathrm{0.8}}$   & 64.14$_{\pm \mathrm{0.82}}$  & 35.30$_{\pm \mathrm{0.59}}$ & 22.48$_{\pm \mathrm{0.4}}$  \\
GNN+WS\cite{Fu2022WaveSANWB}$_{(\mathrm{Arxiv)}}$       & 50.25$_{\pm \mathrm{0.7}}$   & 33.55$_{\pm \mathrm{0.6}}$    & 57.75$_{\pm \mathrm{0.8}}$    & 40.71$_{\pm \mathrm{0.7}}$    & 70.80$_{\pm \mathrm{1.0}}$    & 69.64$_{\pm \mathrm{1.0}}$   & 33.35$_{\pm \mathrm{0.7}}$  & 22.93$_{\pm \mathrm{0.5}}$   \\
FWT+WS\cite{Fu2022WaveSANWB}$_{(\mathrm{Arxiv)}}$      & 50.33$_{\pm \mathrm{0.7}}$   & 32.69$_{\pm \mathrm{0.6}}$    & 57.84$_{\pm \mathrm{0.8}}$    & 38.25$_{\pm \mathrm{0.6}}$    & 69.65$_{\pm \mathrm{1.0}}$    & 65.50$_{\pm \mathrm{1.1}}$   & 33.09$_{\pm \mathrm{0.7}}$  & 22.39$_{\pm \mathrm{0.5}}$   \\
MN+AFA\cite{Hu2022AdversarialFA}$_{(\mathrm{ECCV22)}}$             & 41.02$_{\pm \mathrm{0.4}}$   & 33.52$_{\pm \mathrm{0.4}}$    & 54.66$_{\pm \mathrm{0.5}}$    & 37.60$_{\pm \mathrm{0.4}}$    & 60.71$_{\pm \mathrm{0.5}}$    & 61.28$_{\pm \mathrm{0.5}}$   & 32.32$_{\pm \mathrm{0.3}}$  & 22.11$_{\pm \mathrm{0.2}}$   \\
GNN+AFA\cite{Hu2022AdversarialFA}$_{(\mathrm{ECCV22)}}$           & 46.86$_{\pm \mathrm{0.5}}$   & 34.25$_{\pm \mathrm{0.4}}$    & 54.04$_{\pm \mathrm{0.6}}$    & 36.76$_{\pm \mathrm{0.4}}$   & 67.61$_{\pm \mathrm{0.5}}$    & 63.12$_{\pm \mathrm{0.5}}$   & 33.21$_{\pm \mathrm{0.3}}$  & 22.92$_{\pm \mathrm{0.2}}$   \\
TPN+AFA\cite{Hu2022AdversarialFA}$_{(\mathrm{ECCV22)}}$           & 50.85$_{\pm \mathrm{0.4}}$   & 38.43$_{\pm \mathrm{0.4}}$    & 60.29$_{\pm \mathrm{0.5}}$    & 40.27$_{\pm \mathrm{0.4}}$    & 72.44$_{\pm \mathrm{0.6}}$    & 66.17$_{\pm \mathrm{0.4}}$   & 34.25$_{\pm \mathrm{0.4}}$  & 21.69$_{\pm \mathrm{0.1}}$   \\
RDC\cite{Li2021RankingDC}$_{(\mathrm{CVPR22)}}$                & 48.68$_{\pm \mathrm{0.5}}$   & 38.26$_{\pm \mathrm{0.5}}$    & 59.53$_{\pm \mathrm{0.5}}$    & 42.29$_{\pm \mathrm{0.5}}$    & 79.72$_{\pm \mathrm{0.5}}$    & 65.58$_{\pm \mathrm{0.5}}$   & 32.33$_{\pm \mathrm{0.3}}$  & 22.77$_{\pm \mathrm{0.2}}$   \\
RDC-FT\cite{Li2021RankingDC}$_{(\mathrm{CVPR22)}}$            & 51.20$_{\pm \mathrm{0.5}}$   & 39.13$_{\pm \mathrm{0.5}}$    & 61.50$_{\pm \mathrm{0.6}}$    & 44.33$_{\pm \mathrm{0.6}}$    & 86.33$_{\pm \mathrm{0.5}}$    & 71.57$_{\pm \mathrm{0.5}}$   & 35.84$_{\pm \mathrm{0.4}}$  & 22.27$_{\pm \mathrm{0.2}}$   \\
RN+ATA\cite{Wang2021CrossDomainFC}$_{(\mathrm{IJCAI21)}}$   & 43.02$_{\pm \mathrm{0.4}}$   & 31.79$_{\pm \mathrm{0.3}}$    & 51.16$_{\pm \mathrm{0.5}}$    & 33.72$_{\pm \mathrm{0.3}}$    & 61.17$_{\pm \mathrm{0.5}}$    & 55.69$_{\pm \mathrm{0.5}}$   & 31.13$_{\pm \mathrm{0.3}}$  & 22.14$_{\pm \mathrm{0.2}}$   \\
GNN+ATA\cite{Wang2021CrossDomainFC}$_{(\mathrm{IJCAI21)}}$           & 45.00$_{\pm \mathrm{0.5}}$   & 33.61$_{\pm \mathrm{0.4}}$    & 53.57$_{\pm \mathrm{0.5}}$    & 34.42$_{\pm \mathrm{0.4}}$    & 67.47$_{\pm \mathrm{0.5}}$    & 61.35$_{\pm \mathrm{0.5}}$   & 33.21$_{\pm \mathrm{0.4}}$  & 22.10$_{\pm \mathrm{0.2}}$   \\
TPN+ATA\cite{Wang2021CrossDomainFC}$_{(\mathrm{IJCAI21)}}$           & 50.26$_{\pm \mathrm{0.5}}$   & 34.18$_{\pm \mathrm{0.4}}$    & 57.03$_{\pm \mathrm{0.5}}$    & 39.83$_{\pm \mathrm{0.4}}$    & 77.82$_{\pm \mathrm{0.5}}$    & 65.94$_{\pm \mathrm{0.5}}$   & 34.70$_{\pm \mathrm{0.4}}$  & 21.67$_{\pm \mathrm{0.2}}$   \\
ME-D2N$^{*}$\cite{Fu2022MED2NMD}$_{(\mathrm{ACM \, MM22)}}$            & 65.05$_{\pm \mathrm{0.8}}$   & 49.53$_{\pm \mathrm{0.8}}$    & 60.36$_{\pm \mathrm{0.9}}$    & 52.89$_{\pm \mathrm{0.8}}$    & -            & -           & -          & -           \\
TGDM$^{*}$\cite{Zhuo2022TGDMTG}$_{(\mathrm{ACM \, MM22)}}$              & 64.80$_{\pm \mathrm{0.3}}$   & 50.70$_{\pm \mathrm{0.2}}$    & 61.88$_{\pm \mathrm{0.3}}$    & 52.39$_{\pm \mathrm{0.3}}$    & -            & -           & -          & -           \\
M-FDM$^{*}$\cite{fu2021meta}$_{(\mathrm{ACM \, MM21)}}$      & 63.24$_{\pm \mathrm{0.8}}$   & 51.31$_{\pm \mathrm{0.8}}$    & 58.22$_{\pm \mathrm{0.8}}$    & 51.03$_{\pm \mathrm{0.8}}$    & -            & -           & -          & -           \\
GM-FDM$^{*}$\cite{Fu2022GeneralizedMC}$_{(\mathrm{TIP22)}}$     & 63.85$_{\pm \mathrm{0.4}}$   & 53.10$_{\pm \mathrm{0.4}}$    & 59.39$_{\pm \mathrm{0.4}}$    & 51.28$_{\pm \mathrm{0.4}}$    & 70.21$_{\pm \mathrm{0.4}}$    & 91.07$_{\pm \mathrm{0.4}}$   & 70.90$_{\pm \mathrm{0.6}}$  & 53.07$_{\pm \mathrm{0.4}}$   \\
DDN$^{*}$\cite{Islam2021DynamicDN}$_{(\mathrm{NIPS21)}}$               & -           & -            & -            & -            & 82.14$_{\pm \mathrm{0.8}}$    & 73.14$_{\pm \mathrm{0.8}}$   & 34.66$_{\pm \mathrm{0.6}}$  & 23.38$_{\pm \mathrm{0.4}}$   \\
\bottomrule
\end{tabular}
\label{tab:1_shot_cls}
\end{table}

\begin{table}[h]
\centering
\footnotesize
\tabcolsep=5.5pt
\caption{The CDFS classification result of 5-way k-shot. $^\dag$ denotes the leave-one-out experiments, $^*$ denotes the target data is accessible, $^\mathrm{T}$ denotes the transductive methods.}
\begin{tabular}{l|llllllll}
\toprule
5-way 5-shot&CUB&Cars&Places&Plantae&CropDiseases&EuroSAT&ISIC&ChestX\\
\hline
MatchingNet\cite{Vinyals2016MatchingNF}$_{(\mathrm{NIPS16)}}$&51.37$_{\pm\mathrm{0.8}}$&38.99$_{\pm\mathrm{0.6}}$&63.16$_{\pm\mathrm{0.8}}$&46.53$_{\pm\mathrm{0.9}}$&-&-&-&-\\
MN+FT\cite{Tseng2020CrossDomainFC}$_{(\mathrm{ICLR20)}}$&55.23$_{\pm\mathrm{0.8}}$&41.24$_{\pm\mathrm{0.6}}$&64.55$_{\pm\mathrm{0.8}}$&41.69$_{\pm\mathrm{0.6}}$&-&-&-&-\\
RelationNet\cite{Sung2017LearningTC}$_{(\mathrm{CVPR18)}}$&57.77$_{\pm\mathrm{0.7}}$&37.33$_{\pm\mathrm{0.7}}$&63.32$_{\pm\mathrm{0.8}}$&44.00$_{\pm\mathrm{0.6}}$&-&-&-&-\\
RN+FT\cite{Tseng2020CrossDomainFC}$_{(\mathrm{ICLR20)}}$&59.46$_{\pm\mathrm{0.7}}$&39.91$_{\pm\mathrm{0.7}}$&66.28$_{\pm\mathrm{0.7}}$&45.08$_{\pm\mathrm{0.6}}$&-&-&-&-\\
GNN\cite{Satorras2017FewShotLW}$_{(\mathrm{ICLR18)}}$&62.25$_{\pm\mathrm{0.6}}$&44.28$_{\pm\mathrm{0.6}}$&70.84$_{\pm\mathrm{0.7}}$&52.53$_{\pm\mathrm{0.6}}$&-&-&-&-\\
GNN+FT\cite{Tseng2020CrossDomainFC}$_{(\mathrm{ICLR20)}}$&66.98$_{\pm\mathrm{0.7}}$&44.90$_{\pm\mathrm{0.6}}$&73.94$_{\pm\mathrm{0.7}}$&53.85$_{\pm\mathrm{0.6}}$&-&-&-&-\\
RN+LRP\cite{Sun2020ExplanationGuidedTF}$_{(\mathrm{ICPR20)}}$&59.30$_{\pm\mathrm{0.4}}$&39.19$_{\pm\mathrm{0.4}}$&66.90$_{\pm\mathrm{0.4}}$&48.09$_{\pm\mathrm{0.4}}$&-&-&-&-\\
RN+LRP$^\mathrm{T}$\cite{Sun2020ExplanationGuidedTF}$_{(\mathrm{ICPR20)}}$&59.22$_{\pm\mathrm{0.4}}$&38.31$_{\pm\mathrm{0.4}}$&68.25$_{\pm\mathrm{0.4}}$&47.67$_{\pm\mathrm{0.4}}$&-&-&-&-\\
CAN+LRP\cite{Sun2020ExplanationGuidedTF}$_{(\mathrm{ICPR20)}}$&66.58$_{\pm\mathrm{0.3}}$&43.86$_{\pm\mathrm{0.4}}$&74.91$_{\pm\mathrm{0.4}}$&53.25$_{\pm\mathrm{0.4}}$&-&-&-&-\\
CAN+LRP$^\mathrm{T}$\cite{Sun2020ExplanationGuidedTF}$_{(\mathrm{ICPR20)}}$&66.57$_{\pm\mathrm{0.4}}$&42.57$_{\pm\mathrm{0.4}}$&76.90$_{\pm\mathrm{0.4}}$&51.63$_{\pm\mathrm{0.4}}$&-&-&-&-\\
GNN+LRP\cite{Sun2020ExplanationGuidedTF}$_{(\mathrm{ICPR20)}}$&64.44$_{\pm\mathrm{0.5}}$&46.20$_{\pm\mathrm{0.5}}$&74.45$_{\pm\mathrm{0.5}}$&54.46$_{\pm\mathrm{0.5}}$&-&-&-&-\\
GNN+FT$^\dag$\cite{Tseng2020CrossDomainFC}$_{(\mathrm{ICLR20)}}$&73.11$_{\pm\mathrm{0.7}}$&49.88$_{\pm\mathrm{0.7}}$&77.05$_{\pm\mathrm{0.7}}$&58.84$_{\pm\mathrm{0.7}}$&-&-&-&-\\
CosML$^\dag$\cite{Peng2020CombiningDM}$_{(\mathrm{Arxiv)}}$&66.15$_{\pm\mathrm{0.6}}$&60.17$_{\pm\mathrm{0.6}}$&88.08$_{\pm\mathrm{0.5}}$&42.96$_{\pm\mathrm{0.6}}$&-&-&-&-\\
LRFG$^\dag$\cite{Chen2022CrossDomainFC}$_{(\mathrm{KBS22)}}$&73.94$_{\pm\mathrm{0.7}}$&50.63$_{\pm\mathrm{0.7}}$&76.68$_{\pm\mathrm{0.6}}$&62.14$_{\pm\mathrm{0.7}}$&-&-&-&-\\
DFTL$^\dag$\cite{Yalan2021CrossDomainFC}$_{(\mathrm{ICAICA21)}}$&69.75$_{\pm\mathrm{0.7}}$&49.55$_{\pm\mathrm{0.7}}$&69.38$_{\pm\mathrm{0.7}}$&58.76$_{\pm\mathrm{0.6}}$&-&-&-&-\\
DFTL\cite{Yalan2021CrossDomainFC}$_{(\mathrm{ICAICA21)}}$&69.35$_{\pm\mathrm{0.7}}$&47.91$_{\pm\mathrm{0.6}}$&69.12$_{\pm\mathrm{0.8}}$&58.12$_{\pm\mathrm{0.7}}$&-&-&-&-\\
GNN+MR$^\dag$\cite{Xu2022MemREINRT}$_{(\mathrm{IJCAI22)}}$&77.54$_{\pm\mathrm{0.6}}$&56.78$_{\pm\mathrm{0.6}}$&78.84$_{\pm\mathrm{0.6}}$&65.44$_{\pm\mathrm{0.6}}$&-&-&-&-\\
MNet+MR$^\dag$\cite{Xu2022MemREINRT}$_{(\mathrm{IJCAI22)}}$&67.31$_{\pm\mathrm{0.5}}$&47.36$_{\pm\mathrm{0.5}}$&68.14$_{\pm\mathrm{0.6}}$&52.28$_{\pm\mathrm{0.5}}$&-&-&-&-\\
RNet+MR$^\dag$\cite{Xu2022MemREINRT}$_{(\mathrm{IJCAI22)}}$&68.39$_{\pm\mathrm{0.5}}$&46.92$_{\pm\mathrm{0.5}}$&69.87$_{\pm\mathrm{0.5}}$&58.64$_{\pm\mathrm{0.5}}$&-&-&-&-\\
MAP\cite{Lin2021ModularAF}$_{(\mathrm{Arxiv)}}$&67.92$_{\pm\mathrm{1.1}}$&51.64$_{\pm\mathrm{1.2}}$&75.94$_{\pm\mathrm{1.0}}$&58.45$_{\pm\mathrm{1.2}}$&90.29$_{\pm\mathrm{1.6}}$&82.76$_{\pm\mathrm{2.0}}$&47.85$_{\pm\mathrm{2.0}}$&24.79$_{\pm\mathrm{1.2}}$\\
RN+ST\cite{Liu2021SelfTaughtCF}$_{(\mathrm{Arxiv)}}$&62.94$_{\pm\mathrm{0.4}}$&43.26$_{\pm\mathrm{0.4}}$&66.74$_{\pm\mathrm{0.4}}$&46.92$_{\pm\mathrm{0.3}}$&78.62$_{\pm\mathrm{0.4}}$&75.84$_{\pm\mathrm{0.4}}$&44.42$_{\pm\mathrm{0.3}}$&24.79$_{\pm\mathrm{0.2}}$\\
MN+AFA\cite{Hu2022AdversarialFA}$_{(\mathrm{ECCV22)}}$&59.46$_{\pm\mathrm{0.4}}$&46.13$_{\pm\mathrm{0.4}}$&68.87$_{\pm\mathrm{0.4}}$&52.43$_{\pm\mathrm{0.4}}$&80.07$_{\pm\mathrm{0.4}}$&69.63$_{\pm\mathrm{0.5}}$&39.88$_{\pm\mathrm{0.3}}$&23.18$_{\pm\mathrm{0.2}}$\\
GNN+AFA\cite{Hu2022AdversarialFA}$_{(\mathrm{ECCV22)}}$&68.25$_{\pm\mathrm{0.5}}$&49.28$_{\pm\mathrm{0.5}}$&76.21$_{\pm\mathrm{0.5}}$&54.26$_{\pm\mathrm{0.4}}$&88.06$_{\pm\mathrm{0.3}}$&85.58$_{\pm\mathrm{0.4}}$&46.01$_{\pm\mathrm{0.4}}$&25.02$_{\pm\mathrm{0.2}}$\\
TPN+AFA\cite{Hu2022AdversarialFA}$_{(\mathrm{ECCV22)}}$&65.86$_{\pm\mathrm{0.4}}$&47.89$_{\pm\mathrm{0.4}}$&72.81$_{\pm\mathrm{0.4}}$&55.67$_{\pm\mathrm{0.4}}$&85.69$_{\pm\mathrm{0.4}}$&80.12$_{\pm\mathrm{0.4}}$&46.29$_{\pm\mathrm{0.3}}$&23.47$_{\pm\mathrm{0.2}}$\\
RDC\cite{Li2021RankingDC}$_{(\mathrm{CVPR22)}}$&64.36$_{\pm\mathrm{0.4}}$&52.15$_{\pm\mathrm{0.4}}$&73.24$_{\pm\mathrm{0.4}}$&57.50$_{\pm\mathrm{0.4}}$&88.90$_{\pm\mathrm{0.3}}$&77.15$_{\pm\mathrm{0.4}}$&41.28$_{\pm\mathrm{0.3}}$&25.91$_{\pm\mathrm{0.2}}$\\
RDC-FT\cite{Li2021RankingDC}$_{(\mathrm{CVPR22)}}$&67.77$_{\pm\mathrm{0.4}}$&53.75$_{\pm\mathrm{0.5}}$&74.65$_{\pm\mathrm{0.4}}$&60.63$_{\pm\mathrm{0.4}}$&93.55$_{\pm\mathrm{0.3}}$&84.67$_{\pm\mathrm{0.3}}$&49.06$_{\pm\mathrm{0.3}}$&25.48$_{\pm\mathrm{0.2}}$\\
RN+ATA\cite{Wang2021CrossDomainFC}$_{(\mathrm{IJCAI21)}}$&59.36$_{\pm\mathrm{0.4}}$&42.95$_{\pm\mathrm{0.4}}$&66.90$_{\pm\mathrm{0.4}}$&45.32$_{\pm\mathrm{0.3}}$&78.20$_{\pm\mathrm{0.4}}$&71.02$_{\pm\mathrm{0.4}}$&40.38$_{\pm\mathrm{0.3}}$&24.43$_{\pm\mathrm{0.2}}$\\
GNN+ATA\cite{Wang2021CrossDomainFC}$_{(\mathrm{IJCAI21)}}$&66.22$_{\pm\mathrm{0.5}}$&49.14$_{\pm\mathrm{0.4}}$&75.48$_{\pm\mathrm{0.4}}$&52.69$_{\pm\mathrm{0.4}}$&90.59$_{\pm\mathrm{0.3}}$&83.75$_{\pm\mathrm{0.4}}$&44.91$_{\pm\mathrm{0.4}}$&24.32$_{\pm\mathrm{0.4}}$\\
TPN+ATA\cite{Wang2021CrossDomainFC}$_{(\mathrm{IJCAI21)}}$&65.31$_{\pm\mathrm{0.4}}$&46.95$_{\pm\mathrm{0.4}}$&72.12$_{\pm\mathrm{0.4}}$&55.08$_{\pm\mathrm{0.4}}$&88.15$_{\pm\mathrm{0.5}}$&79.47$_{\pm\mathrm{0.3}}$&45.83$_{\pm\mathrm{0.3}}$&23.60$_{\pm\mathrm{0.2}}$\\
NSAE\cite{Liang2021BoostingTG}$_{(\mathrm{ICCV21)}}$&76.00$_{\pm\mathrm{0.7}}$&61.11$_{\pm\mathrm{0.8}}$&73.40$_{\pm\mathrm{0.7}}$&65.66$_{\pm\mathrm{0.8}}$&96.09$_{\pm\mathrm{0.4}}$&87.53$_{\pm\mathrm{0.5}}$&56.85$_{\pm\mathrm{0.7}}$&28.73$_{\pm\mathrm{0.5}}$\\
TACDFSL\cite{Zhang2022TACDFSLTA}$_{(\mathrm{SB22)}}$&-&-&-&-&93.42$_{\pm\mathrm{0.6}}$&85.19$_{\pm\mathrm{0.7}}$&45.39$_{\pm\mathrm{0.7}}$&25.32$_{\pm\mathrm{0.5}}$\\
TMHFS\cite{Jiang2020ATM}$_{(\mathrm{Arxiv)}}$&-&-&-&-&95.28$_{\pm\mathrm{0.4}}$&85.34$_{\pm\mathrm{0.6}}$&53.84$_{\pm\mathrm{0.7}}$&27.98$_{\pm\mathrm{0.5}}$\\
BSR\cite{Liu2020FeatureTE}$_{(\mathrm{Arxiv)}}$&-&-&-&-&96.59$_{\pm\mathrm{0.3}}$&88.13$_{\pm\mathrm{0.5}}$&57.40$_{\pm\mathrm{0.7}}$&29.72$_{\pm\mathrm{0.5}}$\\
SB-MTL\cite{Cai2020SBMTLSM}$_{(\mathrm{Arxiv)}}$&-&-&-&-&96.01$_{\pm\mathrm{0.4}}$&85.93$_{\pm\mathrm{0.7}}$&50.68$_{\pm\mathrm{0.8}}$&25.99$_{\pm\mathrm{0.5}}$\\
SSP\cite{Zhang2022HowWD}$_{(\mathrm{Arxiv)}}$&-&-&-&-&88.09$_{\pm\mathrm{0.6}}$&81.10$_{\pm\mathrm{0.6}}$&43.74$_{\pm\mathrm{0.6}}$&26.80$_{\pm\mathrm{0.5}}$\\
ReFine\cite{Oh2022ReFineRB}$_{(\mathrm{CIKM22)}}$&-&-&-&-&90.75$_{\pm\mathrm{0.5}}$&82.36$_{\pm\mathrm{0.6}}$&51.68$_{\pm\mathrm{0.6}}$&26.76$_{\pm\mathrm{0.4}}$\\
GNN+WS\cite{Fu2022WaveSANWB}$_{(\mathrm{Arxiv)}}$&70.31$_{\pm\mathrm{0.7}}$&46.11$_{\pm\mathrm{0.7}}$&76.88$_{\pm\mathrm{0.6}}$&57.72$_{\pm\mathrm{0.6}}$&89.70$_{\pm\mathrm{0.6}}$&85.22$_{\pm\mathrm{0.7}}$&44.93$_{\pm\mathrm{0.7}}$&25.63$_{\pm\mathrm{0.5}}$\\
FWT+WS\cite{Fu2022WaveSANWB}$_{(\mathrm{Arxiv)}}$&71.16$_{\pm\mathrm{0.7}}$&47.78$_{\pm\mathrm{0.7}}$&78.19$_{\pm\mathrm{0.6}}$&57.85$_{\pm\mathrm{0.7}}$&91.23$_{\pm\mathrm{0.5}}$&84.84$_{\pm\mathrm{0.7}}$&46.00$_{\pm\mathrm{0.7}}$&25.27$_{\pm\mathrm{0.5}}$\\
ME-D2N$^{*}$\cite{Fu2022MED2NMD}$_{(\mathrm{ACM\,MM22)}}$&83.17$_{\pm\mathrm{0.6}}$&69.17$_{\pm\mathrm{0.7}}$&80.45$_{\pm\mathrm{0.6}}$&72.87$_{\pm\mathrm{0.7}}$&-&-&-&-\\
TGDM$^{*}$\cite{Zhuo2022TGDMTG}$_{(\mathrm{ACM\,MM22)}}$&84.21$_{\pm\mathrm{0.2}}$&70.99$_{\pm\mathrm{0.2}}$&81.62$_{\pm\mathrm{0.2}}$&71.78$_{\pm\mathrm{0.2}}$&-&-&-&-\\
M-FDM$^{*}$\cite{fu2021meta}$_{(\mathrm{ACM\,MM21)}}$&79.46$_{\pm\mathrm{0.6}}$&66.52$_{\pm\mathrm{0.7}}$&78.92$_{\pm\mathrm{0.6}}$&69.22$_{\pm\mathrm{0.7}}$&-&-&-&-\\
GM-FDM$^{*}$\cite{Fu2022GeneralizedMC}$_{(\mathrm{TIP22)}}$&80.49$_{\pm\mathrm{0.3}}$&71.80$_{\pm\mathrm{0.3}}$&78.80$_{\pm\mathrm{0.3}}$&69.45$_{\pm\mathrm{0.3}}$&87.32$_{\pm\mathrm{0.2}}$&95.87$_{\pm\mathrm{0.2}}$&84.07$_{\pm\mathrm{0.4}}$&55.37$_{\pm\mathrm{0.4}}$\\
DDN$^{*}$\cite{Islam2021DynamicDN}$_{(\mathrm{NIPS21)}}$&-&-&-&-&95.54$_{\pm\mathrm{0.4}}$&89.07$_{\pm\mathrm{0.5}}$&49.36$_{\pm\mathrm{0.6}}$&28.31$_{\pm\mathrm{0.5}}$\\
\hline
5-way 20-shot&CUB&Cars&Places&Plantae&CropDiseases&EuroSAT&ISIC&ChestX\\
\hline
BSR\cite{Liu2020FeatureTE}$_{(\mathrm{Arxiv)}}$&-&-&-&-&99.16$_{\pm\mathrm{0.1}}$&94.72$_{\pm\mathrm{0.3}}$&68.09$_{\pm\mathrm{0.6}}$&38.34$_{\pm\mathrm{0.5}}$\\
TMHFS\cite{Jiang2020ATM}$_{(\mathrm{Arxiv)}}$&-&-&-&-&98.51$_{\pm\mathrm{0.2}}$&92.42$_{\pm\mathrm{0.4}}$&65.43$_{\pm\mathrm{0.6}}$&37.11$_{\pm\mathrm{0.5}}$\\
SB-MTL\cite{Cai2020SBMTLSM}$_{(\mathrm{Arxiv)}}$&-&-&-&-&99.19$_{\pm\mathrm{0.1}}$&95.18$_{\pm\mathrm{0.4}}$&68.58$_{\pm\mathrm{0.7}}$&33.47$_{\pm\mathrm{0.5}}$\\
SSP\cite{Zhang2022HowWD}$_{(\mathrm{Arxiv)}}$&-&-&-&-&94.95$_{\pm\mathrm{0.3}}$&88.54$_{\pm\mathrm{0.5}}$&54.61$_{\pm\mathrm{0.5}}$&32.90$_{\pm\mathrm{0.5}}$\\
NSAE\cite{Liang2021BoostingTG}$_{(\mathrm{ICCV21)}}$&91.08$_{\pm\mathrm{0.4}}$&85.04$_{\pm\mathrm{0.5}}$&83.00$_{\pm\mathrm{0.6}}$&81.54$_{\pm\mathrm{0.6}}$&99.20$_{\pm\mathrm{0.1}}$&94.21$_{\pm\mathrm{0.3}}$&67.45$_{\pm\mathrm{0.6}}$&36.14$_{\pm\mathrm{0.5}}$\\
TACDFSL\cite{Zhang2022TACDFSLTA}$_{(\mathrm{SB22)}}$&-&-&-&-&95.49$_{\pm\mathrm{0.4}}$&87.87$_{\pm\mathrm{0.5}}$&53.15$_{\pm\mathrm{0.6}}$&29.17$_{\pm\mathrm{0.5}}$\\
\hline
5-way 50-shot&CUB&Cars&Places&Plantae&CropDiseases&EuroSAT&ISIC&ChestX\\
\hline
BSR\cite{Liu2020FeatureTE}$_{(\mathrm{Arxiv)}}$&-&-&-&-&99.73$_{\pm\mathrm{0.1}}$&96.89$_{\pm\mathrm{0.2}}$&74.08$_{\pm\mathrm{0.6}}$&44.43$_{\pm\mathrm{0.6}}$\\
TMHFS\cite{Jiang2020ATM}$_{(\mathrm{Arxiv)}}$&-&-&-&-&99.28$_{\pm\mathrm{0.1}}$&95.63$_{\pm\mathrm{0.3}}$&71.29$_{\pm\mathrm{0.8}}$&43.43$_{\pm\mathrm{0.7}}$\\
SB-MTL\cite{Cai2020SBMTLSM}$_{(\mathrm{Arxiv)}}$&-&-&-&-&99.75$_{\pm\mathrm{0.1}}$&97.73$_{\pm\mathrm{0.3}}$&75.55$_{\pm\mathrm{0.6}}$&38.37$_{\pm\mathrm{0.6}}$\\
SSP\cite{Zhang2022HowWD}$_{(\mathrm{Arxiv)}}$&-&-&-&-&96.27$_{\pm\mathrm{0.3}}$&91.40$_{\pm\mathrm{0.4}}$&60.86$_{\pm\mathrm{0.5}}$&37.05$_{\pm\mathrm{0.5}}$\\
NSAE\cite{Liang2021BoostingTG}$_{(\mathrm{ICCV21)}}$&95.41$_{\pm\mathrm{0.5}}$&&86.53$_{\pm\mathrm{0.8}}$&85.99$_{\pm\mathrm{0.7}}$&99.70$_{\pm\mathrm{0.1}}$&96.50$_{\pm\mathrm{0.3}}$&73.00$_{\pm\mathrm{0.6}}$&41.80$_{\pm\mathrm{0.7}}$\\
TACDFSL\cite{Zhang2022TACDFSLTA}$_{(\mathrm{SB22)}}$&-&-&-&-&95.88$_{\pm\mathrm{0.4}}$&89.07$_{\pm\mathrm{0.4}}$&56.68$_{\pm\mathrm{0.6}}$&31.75$_{\pm\mathrm{0.5}}$\\
\bottomrule

\end{tabular}
\label{tab:k_shot_cls}
\end{table}
\clearpage

\begin{figure}[h]
    \centering
    \includegraphics[height=0.4\textwidth,trim={0cm 0cm 0cm 0cm}, clip]{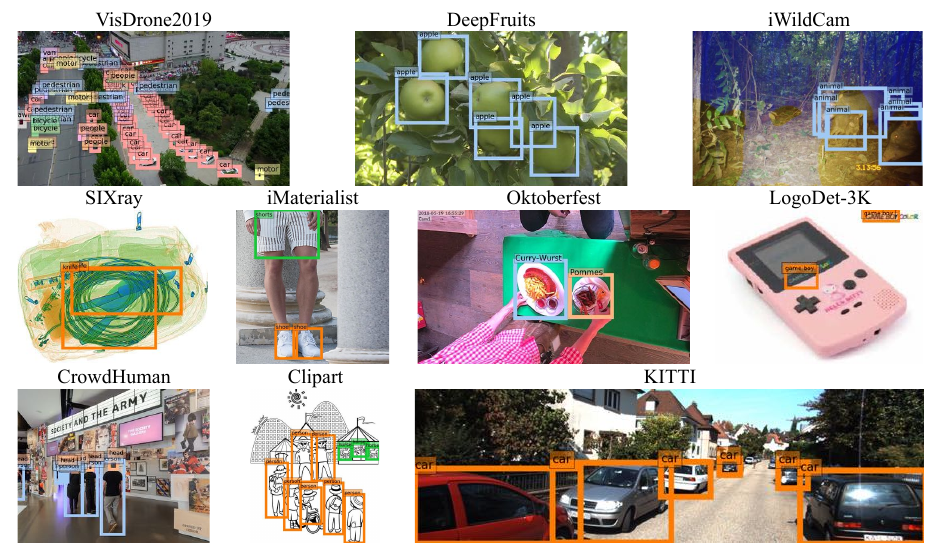}
        \caption{ Sample images in the proposed MoF-SOD \cite{Lee2022RethinkingFO} benchmark.
    }
    \vspace{-0.3cm}
    \label{fig:detection_bm} 
\end{figure}

\textbf{CDFS Object-Detection.} MoF-SOD \cite{Lee2022RethinkingFO} propose a Multi-domain Few-Shot Object Detection benchmark consisting of 10 datasets from a wide range of domains to evaluate few-shot object-detection algorithms, as shown in the Fig. \ref{fig:detection_bm}. Empirical results show several keys
actors that have yet to be explored in previous works.
 Under the proposed benchmark, MoF-SOD \cite{Lee2022RethinkingFO} conducted extensive experiments on the
impact of freezing parameters, different architectures, and different pre-training
datasets.

\textbf{CDFS Segmentation.} PATNet \cite{Lei2022CrossDomainFS} proposes a cross-domain few-shot segmentation benchmark.
 Evaluate both representative few-shot segmentation and transfer learning-based methods on the proposed benchmark.
 And propose a novel Pyramid-Anchor-Transformation based few-shot segmentation
network, in which domain-specific features are transformed into domain-agnostic ones for downstream segmentation modules to fast adapt to unseen domains.
And RD \cite{Wang2022RememberTD} also proposes a cross-domain few-shot segmentation task using different public datasets to validate the model effect. As the most famous datasets, we summarize the segmentation transfer effect from COCO-$2-^{i}$ to PASCAL-$5^{i}$ as shown in Tab. \ref{tab:seg_result}.

\begin{table}[h]
 \footnotesize
 \caption{ Cross-domain few-shot semantic segmentation results on COCO-$20^i$ to PASCAL-$5^i$ task.}
  \resizebox{\linewidth}{!}{%
\begin{tabular}{llcccccccccc}
\toprule
\multicolumn{12}{c}{COCO-$20^i$ to PASCAL-$5^i$} \\ 
\toprule
\multicolumn{1}{l|}{\multirow{2}{*}{Backbone}} &
  \multicolumn{1}{l|}{\multirow{2}{*}{Method}} &
  \multicolumn{5}{c|}{1-shot} &
  \multicolumn{5}{c}{5-shot} \\ \cline{3-12} 
\multicolumn{1}{l|}{} &
  \multicolumn{1}{l|}{} &
  split0 &
  split1 &
  split2 &
  split3 &
  \multicolumn{1}{c|}{Mean} &
  split0 &
  split1 &
  split2 &
  split3 &
  Mean \\
\midrule
\multicolumn{1}{l|}{\multirow{7}{*}{ResNet50}} &
  \multicolumn{1}{l|}{RPMMs\cite{Yang2020PrototypeMM}$_{(\mathrm{ECCV20)}}$} &
  36.3 &
  55.0 &
  52.5 &
  54.6 &
  \multicolumn{1}{c|}{49.6} &
  40.2 &
  58.0 &
  55.2 &
  61.8 &
  53.8 \\
\multicolumn{1}{l|}{} &
  \multicolumn{1}{l|}{RePRI\cite{Boudiaf2021FewShotSW}$_{(\mathrm{CVPR21)}}$} &
  52.4 &
  \textbf{64.3} &
  65.3 &
  71.5 &
  \multicolumn{1}{c|}{63.3} &
  57.0 &
  68.0 &
  70.4 &
  76.2 &
  67.9 \\
\multicolumn{1}{l|}{} &
  \multicolumn{1}{l|}{ASGNet\cite{Li2021AdaptivePL}$_{(\mathrm{CVPR21)}}$} &
  42.5 &
  58.7 &
  65.5 &
  63.0 &
  \multicolumn{1}{c|}{57.4} &
  53.7 &
  \textbf{69.8} &
  67.1 &
  75.9 &
  66.6 \\
\multicolumn{1}{l|}{} &
  \multicolumn{1}{l|}{PFENet\cite{Tian2020PriorGF}$_{(\mathrm{TPAMI)}}$} &
   -&
   -&
   -&
   -&
  \multicolumn{1}{c|}{60.8} &
   -&
   -&
   -&
   -&
  61.9 \\
\multicolumn{1}{l|}{} &
  \multicolumn{1}{l|}{CWT\cite{lu2021SimplerIB}$_{(\mathrm{ICCV21)}}$} &
  53.5 &
  59.2 &
  60.2 &
  64.9 &
  \multicolumn{1}{c|}{59.4} &
  60.3 &
  65.8 &
  67.1 &
  72.8 &
  66.5 \\
\multicolumn{1}{l|}{} &
  \multicolumn{1}{l|}{HSNet\cite{Min2021HypercorrelationSF}$_{(\mathrm{ICCV21)}}$} &
  48.7 &
  61.5 &
  63.0 &
  72.8 &
  \multicolumn{1}{c|}{61.5} &
  58.2 &
  65.9 &
  \textbf{71.8} &
  \textbf{77.9} &
  68.4 \\
\multicolumn{1}{l|}{} &
  \multicolumn{1}{l|}{RD\cite{Wang2022RememberTD}$_{(\mathrm{CVPR22)}}$} &
  \textbf{57.4} &
  62.2 &
  \textbf{68.0} &
  \textbf{74.8} &
  \multicolumn{1}{c|}{\textbf{65.6}} &
  \textbf{65.7} &
  69.2 &
  70.8 &
  75.0 &
  \textbf{70.1} \\
\bottomrule
\multicolumn{1}{l|}{\multirow{3}{*}{ResNet101}} &
  \multicolumn{1}{l|}{SCL\cite{Zhang2021SelfGuidedAC}$_{(\mathrm{CVPR21)}}$} &
  43.1 &
  60.3 &
  66.1 &
  68.1 &
  \multicolumn{1}{c|}{59.4} &
  43.3 &
  61.2 &
  66.5 &
  70.4 &
  60.3 \\
\multicolumn{1}{l|}{} &
  \multicolumn{1}{l|}{HSNet\cite{Min2021HypercorrelationSF}$_{(\mathrm{ICCV21)}}$} &
  46.3 &
  \textbf{64.7} &
  67.7 &
  \textbf{74.2} &
  \multicolumn{1}{c|}{63.2} &
  59.1 &
  69.0 &
  \textbf{73.4} &
  78.7 &
  70.0 \\
\multicolumn{1}{l|}{} &
  \multicolumn{1}{l|}{RD\cite{Wang2022RememberTD}$_{(\mathrm{CVPR22)}}$} &
  \textbf{59.4} &
  64.3 &
  \textbf{70.8} &
  72.0 &
  \multicolumn{1}{c|}{\textbf{66.6}} &
  \textbf{67.2} &
  \textbf{72.7} &
  72.0 &
  \textbf{78.9} &
  \textbf{72.7} \\
\bottomrule
\end{tabular}%
}
    
  	\label{tab:seg_result}
  \end{table}

\section{Application OF CDFS}\label{sec:Application}
\begin{figure}[h]
    \centering
    \includegraphics[height=0.4\textwidth,trim={0cm 0cm 0cm 0cm}, clip]{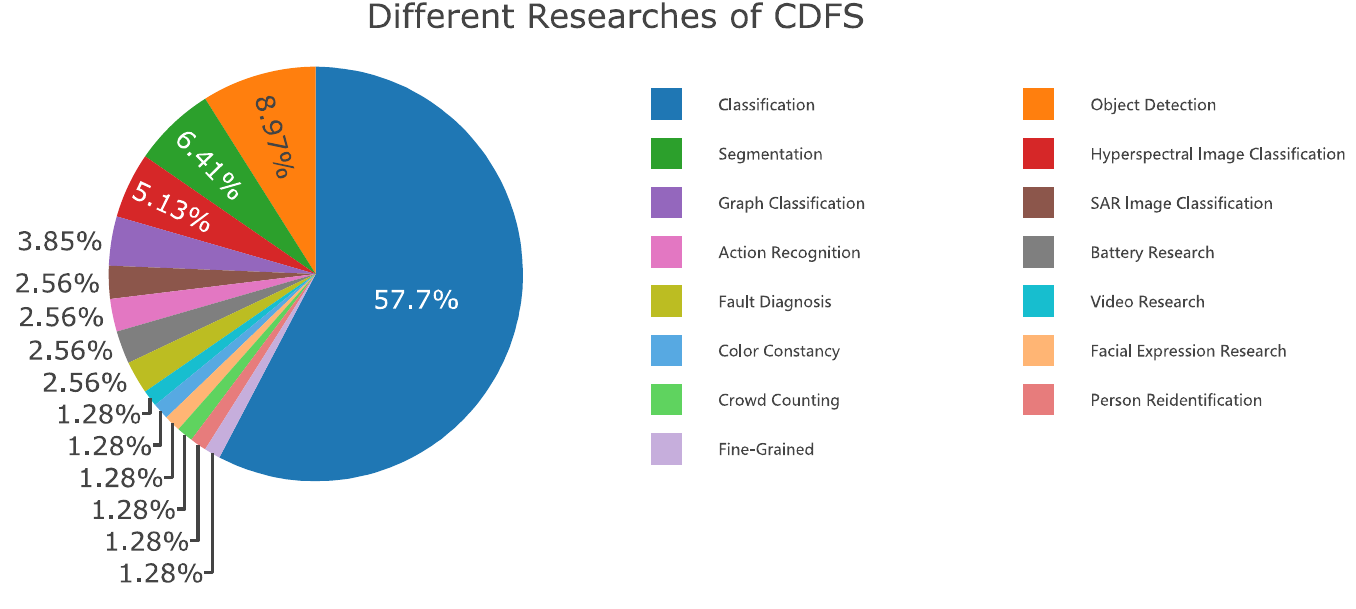}
        \caption{The Cross-Domain Few-shot related applications.
    }
    \vspace{-0.3cm}
    \label{fig:diff_app} 
\end{figure}

The CDFS focuses on the domain problem of few-shot. It has been used in various applications, as shown in Fig. \ref{fig:diff_app}.
We summary the different CDFS methods for object detection and segmentation.
For the other applications, we also provide a detailed survey.

\subsection{CDFS for Object Detection}
\begin{figure}[h]
		\centering
	\includegraphics[height=0.3\textwidth,trim={0cm 0cm 0cm 0cm}, clip]{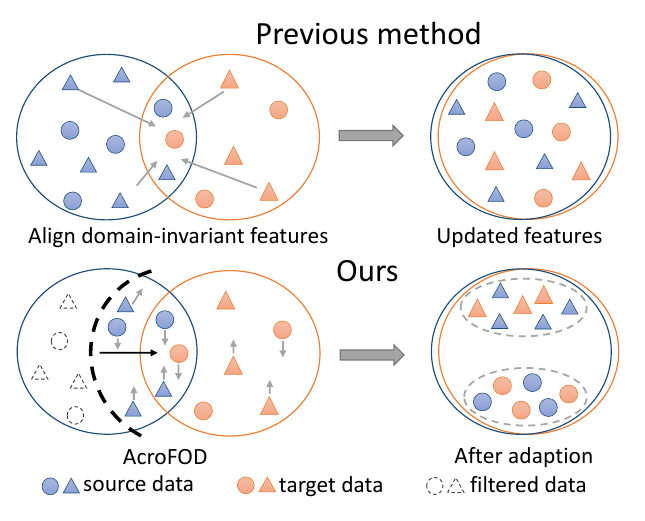}
    		\caption{AcroFOD \cite{Gao2022AcroFODAA} address the task of cross-domain few-shot object detection. Top: Existing
feature-aligning based methods fail to extract discriminative features within limited
labeled data in the target domain. Bottom: Our method filters (thick black dotted
line) source data that is far away from the target domain.
		}
		\vspace{-0.3cm}
		\label{fig:AcroFOD_i}
	\end{figure}
For CDFS object detection, 
 image enhancement is still the most direct choice. CutMix \cite{Nakamura2022FewshotAO} directly pastes data from the target domain to enhance feature diversity.
However, due to the scarcity of target domain data, direct image mixing will bring unnecessary noise impact. To alleviate this situation, AcroFOD \cite{Gao2022AcroFODAA} proposes a filtering mechanism to filter the enhanced image, as shown in Fig. \ref{fig:AcroFOD_i}. In the training process, the model continuously measures the distance between the enhanced and target features, retaining the most similar top-k. Finally, the enhanced features are used for object detection learning.
The generation strategy can still enrich the feature distribution. For this, OA-FSUI2IT \cite{Zhao2022OAFSUI2ITAN} uses the unlabeled data of the target domain to generate new style images for model training. The style of the generated images is consistent with the target domain, and the content is consistent with the source domain. 
Domain adaptive training ensures the reproducibility of the learned content, improves the interpretability of the model, and self-supervised content consistency improves the model generalization.
CDTL \cite{Wang2022FewShotSS} is one of the first to study the problem of few-shot SAR image ship detection, which has great practical value but is less studied than the classification problem of SAR images under few-shot conditions.
CDTL \cite{Wang2022FewShotSS} design a two-stage few-shot SAR ship detection model based on transfer learning using the similarity information of images of optical ships.

\subsection{CDFS for Segmentation}
For CDFS segmentation, RD \cite{Wang2022RememberTD} proposes a domain enhancement strategy of memory mechanism. During training, source domain data continuously stores domain-style information in memory. During testing, source information stored in memory is loaded into target domain feature enhancement. RD \cite{Wang2022RememberTD} can directly reduce the domain difference and has been verified on typically partitioned datasets.
For the task of semantic segmentation in autonomous driving applications,  PixDA \cite{Tavera2021PixelbyPixelCA} propose a novel pixel-by-pixel domain adversarial loss following three criteria: (i) align the source and
the target domain for each pixel, (ii) avoid negative transfer on the correctly represented pixels, and (iii) regularize the training of infrequent classes to avoid overfitting.
CDTF \cite{Lu2022CrossdomainFS} through aligning the support and query prototypes to achieve the cross-domain few-shot segmentation. By aligning query and support
prototypes with an uncertainty-aware contrastive loss, and using a supervised cross-entropy loss and an unsupervised boundary loss as regularizations, CDTF \cite{Lu2022CrossdomainFS} could generalize the base model to the target domain without additional labels. 
 CD-FSS \cite{Wang2022CrossDomainFL} propose a cross-domain few-shot segmentation framework, which enables the model to leverage the learning ability obtained from the natural domain to facilitate rare-disease skin lesion segmentation with limited data of common diseases.

\subsection{CDFS for Other Application}
Cross-Domain Few-shot Hyperspectral Image Classification \cite{Zhang2022GraphIA,Li2021DeepCF,Zhang2022DualGC,Wang2022SpatialSpectralLD} draw the research attention recently, the model focus on learning of local and global image context due to the special image data.
The related hyperspectral framework is similar to the traditional few-shot research.
 SAR-FS \cite{Rostami2019SARIC} developed an algorithm to transfer knowledge from EO domains to SAR domains to eliminate the need for huge labeled data points in the SAR domains. 

 Researchers \cite{Tang2022ANL,Feng2021MetalearningAA}  carried out a series of research and discussion on fault diagnosis, focusing on the impact of meta-learning on cross-domain fault diagnosis.
 FSPR \cite{Dai2022AnEW} first formulated a reidentification scenario as a cross-domain few-shot problem and discussed the difference between conventional and unsupervised re-ID. And introduce a reweighting instance method based on the function (ReWIF) to guide the training procedure of the re-ID model.
 In other related fields \cite{Hou2022SampleEfficientCW,Sun2021AnomalyCA,Zhou2020GoogleHY,Wu2022CrossDomainFL}, CDFS has gradually begun to receive attention, promoting the development of various research subfields.

 \section{FUTURE DIRECTION}\label{sec:FUTURE}
  \subsection{Future Challenges}
CDFS focuses on learning in different domains and largely solves the domain shift challenge. With limited labeled data, current research focuses on near-domain transfer, neglecting the more challenging distant-domain transfer. Transference of models from a natural scene to a proprietary domain remains difficult. For CDFS of multiple sources, the goal of the devised learning scheme is to achieve Domain Adaptation and Generalization. Even the lack of sufficient labeled data in the new CDFS domain may inhibit the transfer effect, reducing the effectiveness of multiple sources in this domain. Thus, fine-tuning variably distributed data among multiple sources is essential to maximize the transfer effect and ensure successful CDFS.
 \subsection{Future Techniques}
Recent researchers have sought to use image and feature enhancement to address the CDFS problem, which requires a model to adapt quickly to a new environment while retaining rapid learning ability. However, successfully combining the insufficient annotation and domain shift issue continues to be challenging. To further develop solutions to the CDFS problem, researchers should consider the importance of adapting high parameters and utilizing domain-invariant information.
\subsubsection{Multiple Source Meta-Learning}
In the future, meta-learning is poised to be a key tool for addressing domain shifts, especially in the context of few-shot learning. The model can successfully manage complex scenarios by leveraging its high adaptability, especially regarding domain migration. As such, upcoming work in cross-domain few-shot (CDFS) should focus more on meta-learning in a multi-source setting \cite{Qiu2021MetaSF}. Exploiting the abundant data distributions available should maximize the utility of meta-learning and preserve its original intention. Additionally, research should be conducted to combine memory mechanisms with meta-learning; such a union of the two would bring straightforward benefits to the CDFS domain. Memory can be key in connecting domains without compromising performance; meanwhile, meta-learning provides highly customizable parameters.
\subsubsection{Domain invariant information}
When faced with single-domain CDFS, domain-invariant information is invaluable for successful zero-shot learning \cite{Min2020DomainAwareVB, Min2019DomainSpecificEN}. Specifically, semantic information can often be used to evidence relationships between different categories. Because visual differences do not hamper it, they can then be readily employed as supplementary information for learning. In the context of FSL, several studies have explored the intersection of semantic modalities and vision, often using the former to facilitate the knowledge migration between categories \cite{Wang2020LargeScaleFL, Peng2019FewShotIR}. When shared category information can be successfully converted into a different modality, it can offer valuable guidance in cross-domain FSL scenarios \cite{Guan2020LargeScaleCF}.
 \subsection{Future Applications}
\subsubsection{Open-Set Few-Shot Learning}
The softmax of the deep model assigns all samples in the test phase to a fixed category, which is different from actual expectations as it cannot identify unseen classes \cite{Salehi2021AUS}. To address this limitation, open-set models are more suitable compared with close-set models as they require the model to make a corresponding response to the unseen class rather than a mechanical one. Cross Domain Few-Shot (CDFS) transfer of the model to the target domain allows for identifying unseen classes, albeit requiring a few fine-tuning annotations when doing so \cite{Liu2020FewShotOR,Jeong2021FewshotOR}. However, this approach is only effective when classes in the source and target domains can be delineated, raising an important question about identifying these unlabeled classes for complex scenarios where all classes need to be labeled. Open-set few-shot learning can help to meet this challenge.
\subsubsection{Incremental Few-Shot Learning}
Incremental few-shot learning requires the model to recognize new classes without forgetting the learned classes \cite{Tao2020FewShotCL,Zhu2021SelfPromotedPR}. Typical few-shot models only focus on new classes, and base classes must be relearned to complete recognition. Data dependency results in the models being effective in limited scenarios and cannot increase classes. 
For future CDFS, the model will constantly learn new domain information during the generalization process. In order to identify the different domain classes, the model needs to learn and save the data of the corresponding domain constantly. Binding data to the model anytime does not conform to the original intention of few-shot learning.
Therefore, studying incremental few-shot learning can promote CDFS to a certain extent. If we can eliminate the dependence on data domains, CDFS will be more practical
\cite{Ganea2021IncrementalFI,Nguyen2022iFSRCNNAI}.

\section{Conclusion}
We have provided a comprehensive overview of recent cross-domain few-shot learning (CDFS) research. Our analysis considered existing solutions and research issues while comparing performance indices between different studies. Furthermore, we discussed the broad applications of CDFS and its implications for future research. Our review will serve as a valuable reference guide and provide theoretical support for advancing the field of CDFS.
Our survey reveals that cross-domain few-shot learning is gradually becoming an increasingly popular research topic and has received extensive attention due to its potential to alleviate the domain shift problem in AI applications. The current solutions to the problem span various approaches, each with its advantages and limitations. As the research field is still in its infancy, future work should focus on extending existing or new methods to improve the performance of cross-domain few-shot learning systems.

\section{Acknowledgements}
This work was supported in part by the National Natural Science Foundation of China (No.62176009), the Major Project for New Generation of AI (No.2018AAA0100400), the National Natural Science Foundation of China (No. 61836014, No. U21B2042, No. 62072457, No. 62006231).
\clearpage


\bibliographystyle{elsarticle-num}

\bibliography{cas-refs}
\clearpage

\bio{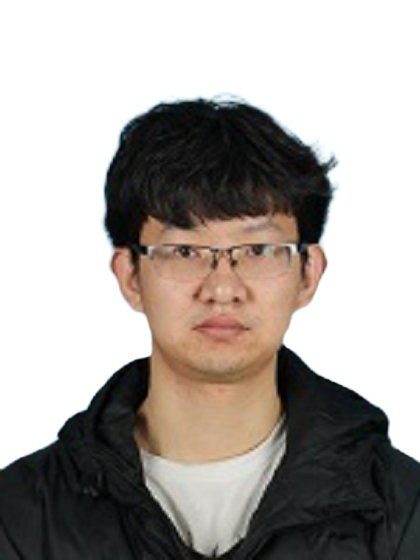}
\textbf{Wenjian Wang} Wenjian Wang received an M.S degree in computer science
from North University of China. He is currently studying for a Ph.D. degree at the  Faculty of Information Technology, Beijing University of 
Technology, China. His current research interests include
Image processing, machine learning, and computer vision.
\\
\\
\\
\\
\\
\endbio

\bio{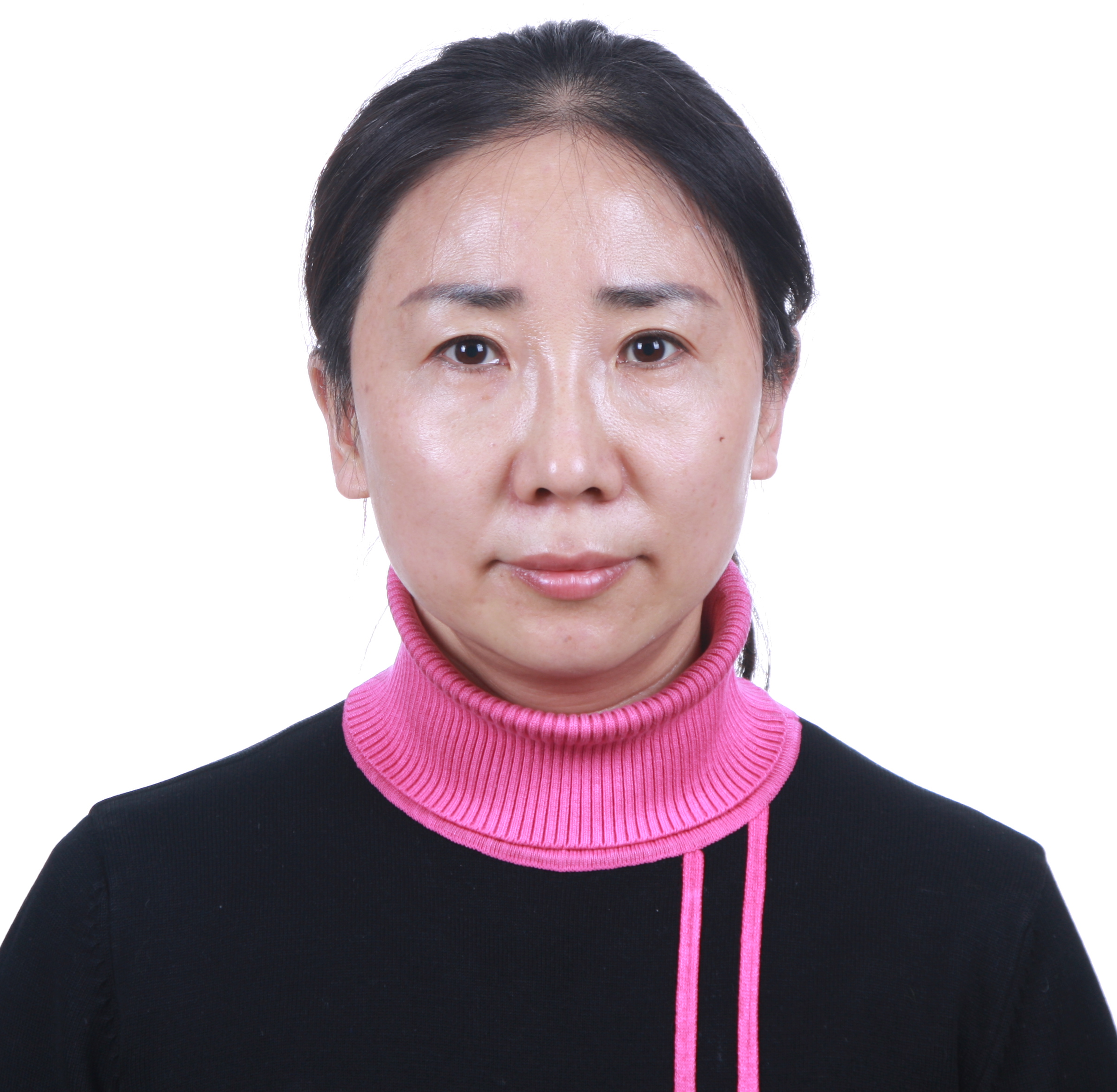}
\textbf{Lijuan Duan} received the B.Sc. and M.Sc. degrees in computer science from Zhengzhou University of Technology, Zhengzhou, China, in 1995 and 1998, respectively. She received the Ph.D. degree in computer science from the Institute of Computing Technology, Chinese Academy of Sciences, Beijing, in 2003. She is currently a Professor at the Faculty of Information Technology, Beijing University of Technology, China. Her research interests include Artificial Intelligence, Image Processing, Computer Vision and Information Security. She has published more than 70 research articles in refereed journals and proceedings on artificial intelligent, image processing and computer vision.

\endbio

\bio{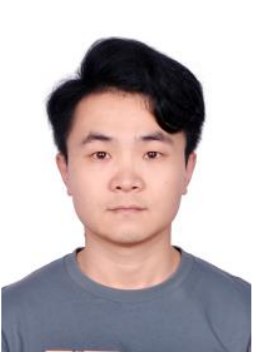}\textbf{Yuxi Wang}
received the Bachelor degree from North Eastern University, China, in 2016, and the PhD degree from University of Chinese Academy of Sciences (UCAS), Institute of Automation, Chinese Academy of Sciences (CASIA), in January 2022. He is now an assistant professor in the Center for Artificial Intelligence and Robotics (CAIR), Hong Kong Institute of Science \& Innovation, Chinese Academy of Science (HKISI-CAS). His research interests include transfer learning, domain adaptation and computer vision.
\\
\\
\\
\endbio

\bio{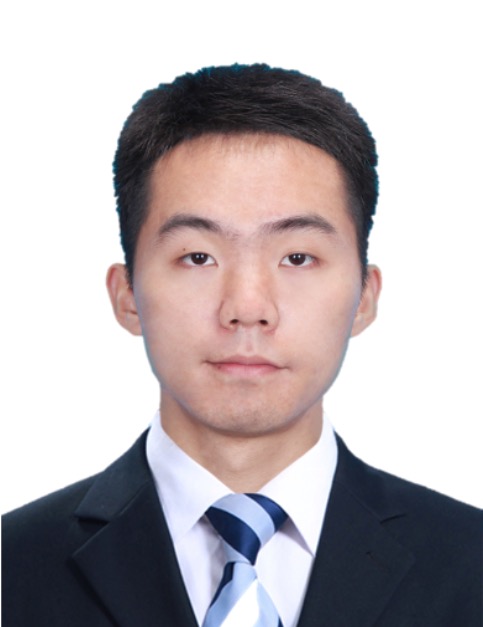}\textbf{Junsong Fan}
 received his bachelor's degree from the School of Automation Science and Electrical Engineering, Beihang University, in 2016. Then, he joined the Center of Research on Intelligent Perception and Computing, National Laboratory of Pattern Recognition, Institute of Automation, Chinese Academy of Sciences, under the supervision of Professor Tieniu Tan and Zhaoxiang Zhang, and received his Ph.D. degree in 2022. He is now an assistant professor in the Centre for Artificial Intelligence and Robotics, Hong Kong Institute of Science \& Innovation, Chinese Academy of Sciences. His research interests include computer vision and machine learning, label-free learning, and visual perception in open worlds.
 \\

\endbio

\bio{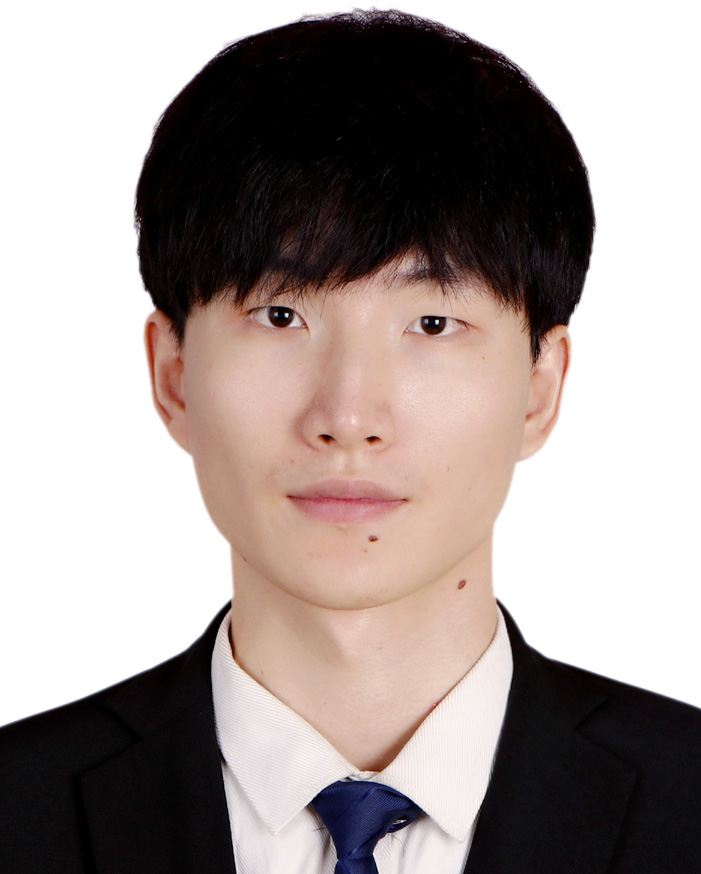}\textbf{Zhi Gong}
Zhi Gong (Student Member,IEEE) received the B.E. degree incomputer science and technology from Shandong Normal University,Shandong,China,in2019. He is currently pursuing the Ph.D. degree in computer science and technology from the Beijing University of Technology,Beijing,China.
His research interests include computer vision and image processing.
 \\
\\
\\
\\
\endbio

\bio{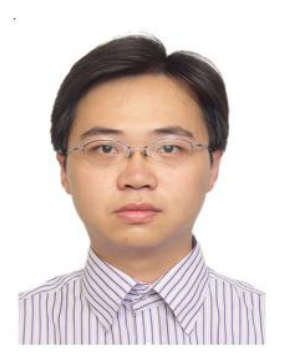}\textbf{Zhaoxiang Zhang} received his bachelor’s degree from the Department of Electronic Science and Technology in the University of Science and Technology of China (USTC) in 2004. After that, he was a Ph.D. candidate under the supervision of Professor Tieniu Tan in the National Laboratory of Pattern Recognition, Institute of Automation, Chinese Academy of Sciences, where he received his Ph.D. degree in 2009. In October 2009, he joined the School of Computer Science and Engineering, Beihang University, as an Assistant Professor (2009-2011), an Associate Professor (2012-2015) and the Vise-Director of the Department of Computer application technology (2014-2015). In July 2015, he returned to the Institute of Automation, Chinese Academy of Sciences. He is now a full Professor in the Center for Research on Intelligent Perception and Computing (CRIPAC) and the National Laboratory of Pattern Recognition (NLPR).
His research interests include Computer Vision, Pattern Recognition, and Machine Learning. Recently, he specifically focuses on deep learning models, biologically-inspired visual computing and human-like learning, and their applications on human analysis and scene understanding.
He has published more than 200 papers in international journals and conferences, including reputable international journals such as JMLR, IEEE TIP, IEEE TNN, IEEE TCSVT, IEEE TIFS and top level international conferences like CVPR, ICCV, NIPS, ECCV, AAAI, IJCAI and ACM MM.
He is serving or has served as the Associated Editor of IEEE T-CSVT, Patten Recognition, Neurocomputing, and Frontiers of Computer Science. He has served as the Area Chair, Senior PC of international conferences like CVPR, NIPS, ICML, AAAI, IJCAI and ACM MM. He is a Senior Member of IEEE, a Distinguished Member of CCF, and a Distinguished member of CAAI.
\\
\\
\\
\endbio

\end{document}